# Modeling, Simulation, and Application of Spatio-Temporal Characteristics Detection in Incipient Slip

Mingxuan Li, *Graduate Student Member, IEEE*, Lunwei Zhang, *Graduate Student Member, IEEE*, Qiyin Huang, Tiemin Li, and Yao Jiang, *Member, IEEE*

*Abstract*—Incipient slip detection provides critical feedback for robotic grasping and manipulation tasks. However, maintaining its adaptability under diverse object properties and complex working conditions remains challenging. This article highlights the importance of completely representing spatio-temporal features of slip, and proposes a novel approach for incipient slip modeling and detection. Based on the analysis of localized displacement phenomenon, we establish the relationship between the characteristic strain rate extreme events and the local slip state. This approach enables the detection of both the spatial distribution and temporal dynamics of stick-slip regions. Also, the proposed method can be applied to strain distribution sensing devices, such as vision-based tactile sensors. Simulations and prototype experiments validated the effectiveness of this approach under varying contact conditions, including different contact geometries, friction coefficients, and combined loads. Experiments demonstrated that this method not only accurately and reliably delineates incipient slip, but also facilitates friction parameter estimation and adaptive grasping control.

## I. INTRODUCTION

TACTILE perception plays a crucial role in stable grasping and dexterous manipulation in humans [1]. Neuroscientific studies show that humans can identify the frictional parameters of objects they touch with over 90% accuracy [2], and quickly adjust the grasp force within about 200 milliseconds to prevent slipping [3]. This ability enables humans to adapt to changes in friction levels based on tactile feedback and apply proper force to ensure stability while maintaining gentle grasping [4].

The perception of incipient slip is an effective means for friction parameter recognition and grasp force control [5],[6]. Incipient slip is an intermediate state between complete sticking and full slipping of the contact surface, as shown in Fig. 1. When a tangential load is applied to the contact surface, slip first occurs at the contact edge. It gradually spreads inward, eventually covering the entire stick region [7]. Besides, at each point on the contact surface, the localized slip state can be described by a time-dependent binary function. We refer to these two characteristics of incipient slip as spatial and temporal characteristics: spatial characteristics refer to the distribution of the stick-slip region at a given moment, while temporal characteristics describe the time evolution of local slip. These characteristics are widely present in human tactile perception. According to existing research, Human sensory information is encoded by neural populations to capture spatial distribution, rather than being transmitted by individual neurons. [8]. Besides, skin deformation can be influenced by the loading history [9]. Due to the time-dependent nature of slip, it is impossible to capture its continuous variation using only instantaneous information. Thus, tactile perception may encode the spatio-temporal features of slip to achieve a comprehensive representation of friction characteristics and contact behavior [10].

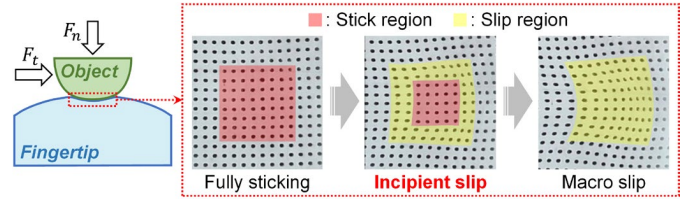

**Fig. 1.** Incipient slip phenomenon. With the increase of tangential load, the slip region gradually expands from the outside to the inside until the stick region disappears. At this time, the contact surface reaches macro slip.

In recent years, studies have focused on slip detection in robots to enhance grasping and manipulation [11]-[15]. To achieve human-like abilities of friction recognition and grasp control in unstructured environments, robots need to tackle challenges posed by unknown object properties and complex loads [16]. To this end, incipient slip detection requires high adaptability in two aspects:

1) *Object adaptability*: the applicability to objects of varying shapes, textures, and stiffness;
2) *Working condition adaptability*: the ability to handle complex and rapidly changing loads and motions during different tasks.

To meet these requirements, robots need to go beyond merely detecting the occurrence of slip. Spatial characteristics reflect the local distribution of contact geometry and friction properties, influencing the stability assessment of objects with different shapes and materials. Temporal characteristics

This work was supported by the National Natural Science Foundation of China under Grant 52375017.

Mingxuan Li, Lunwei Zhang, Qiyin Huang, Tiemin Li, and Yao Jiang are with the Department of Mechanical Engineering, Tsinghua University, Beijing 100084, China (e-mail: mingxuan-li@foxmail.com; zlw21@mails.tsinghua.edu.cn; huangqy21@mails.tsinghua.edu.cn; litm@mails.tsinghua.edu.cn; jiangyao@mail.tsinghua.edu.cn).

describe the dynamic evolution of slip, which is closely related to changes in external loads and motions. Neglecting either side could hinder the ability of stable grasping and dexterous manipulation in robots [17]. Thus, detecting and modeling the spatio-temporal characteristics of incipient slip is necessary for comprehensively analyzing the spatial configuration and temporal evolution of interactions.

Existing incipient slip detection methods can be divided into two categories: time characteristic-based detection and spatial characteristic-based detection.

*1) Detection based on temporal characteristics*: By indirectly characterizing the evolution of incipient slip through lower-dimensional features, the sensor is able to present the temporal properties in terms of the time series changes of these feature values. Song et al. predicted the occurrence of slip through the ratio of friction force to normal force [12]. Wang et al. used deep neural networks in a PapillArray sensor to learn the tangential velocity drop of the pillars to achieve slip detection [18]. Yu et al. used empirical mode decomposition (EMD) based on a floating electrode tactile sensor array to detect the incipient slip [19]. Zhang et al. characterized the localized displacement phenomenon of incipient slip by the distance between the latest slip point and the center point of the contact region [20]. However, the indirect features obtained through dimensionality reduction cannot fully capture the distribution of the stick-slip state. Thus, it is difficult to distinguish the evolution of incipient slip under different contact states (e.g., differentiating the contributions of rotational and translational slip).

*2) Detection based on spatial characteristics*: Recently, vision-based tactile sensors have provided a promising solution for high-resolution measurement [21]. The marker displacement method used in such sensors can characterize contact deformation during the slippage, and provide a foundation for quantitatively detecting the dense spatial characteristics of incipient slip [22]. Dong et al. used the deviation between the deformation and the rigid displacement of the GelSlim sensor's gel to measure incipient slip [23]. Random forest classifier (RFC) [24] and convolutional neural network (CNN) [25] were used to characterize the extent of incipient slip from tactile images containing marker displacements. The Tac3D sensor detected translational incipient slip through deformation field gradients [26], [27], and also measured rotational incipient slip by examining the rotation of line features [28]. However, related methods estimated the degree of incipient slip based on the tactile characteristics at the current moment, without considering the influence of loading history (such as residual strain) on contact deformation. Research indicates that such approaches have limitations when handling complex loading conditions, such as changes in the direction of tangential force and rapid loading conditions [28].

Furthermore, studies have attempted to fully characterize the spatio-temporal information of incipient slip through data-driven approaches [29], [30]. Training networks including both temporal and spatial layers were used. They took several frames of tactile image sequences before the slip occurred as input and outputted the prediction of whether slip would happen. Such strategies captured the inherent time dependence of slippage and compressed high-dimensional tactile images into dynamic features. Thus, both the spatial and temporal dimensions of incipient slip were considered. However, end-to-end mechanisms also lack a complete representation of distribution characteristics, leading to limitations in the output (e.g., only determining the occurrence of slip). Additionally, the dependence on training data restricts the model's ability to generalize to different objects and working conditions.

This article focuses on the modeling, simulation, and application of incipient slip detection. Based on the analysis of contact mechanics models, we use strain-rate extreme events as direct indicators of localized slip. Accordingly, we propose an incipient slip detection method suitable for tactile sensors that can measure strain distribution, such as vision-based tactile sensors. This method is characterized by its ability to comprehensively represent the spatio-temporal features of incipient slip, including the spatial distribution and evolution of local stick-slip states across the entire contact surface. The effectiveness of the proposed method has been evaluated through finite element simulations and prototype experiments. It outperforms existing early slip detection methods in terms of applicability, especially when dealing with complex and rapidly changing loading conditions. Furthermore, the proposed method can be used to measure the stochastic friction distribution that occurs during local slip in the grasping process. When combined with slip- and friction-based control strategies, it can be applied to grasping and manipulation control. Through experiments, we demonstrate the preliminary application of this method in material friction recognition and grasp force control.

The main contributions can be summarized as follows:

1) We constructed a contact mechanics model for the spatio-temporal characteristics of incipient slip and introduced a measurement metric based on strain-rate extreme events.

2) We proposed a novel incipient slip measurement method that achieves comprehensive characterization of the spatio-temporal features of incipient slip. Experiments demonstrate its good adaptability to object properties and contact conditions.

3) Through recognition, grasping, and manipulation tasks, we showcased the application of this method in friction identification and force regulation.

The remainder of this article is organized as follows. Section II provides descriptions of the contact mechanics model and detection method for the spatio-temporal characteristics of incipient slip. Section III details a comprehensive simulation. Section IV presents static experimental evaluations and online grasping assessments. Section V demonstrates applications in friction recognition and grasp force control. Finally, Section VI concludes this article.

## II. Methodology

### A. Spatio-Temporal Characteristics of Incipient Slip

The mechanical mechanism of slip is the relative motion between materials, which is directly manifested as variations in contact stress and strain [31], [32]. Physiological studies on the dynamics of human fingerpads have revealed that afferent nerves encode the strain-rate changes on the surface caused by slippage [10], [33], [34], which are used for incipient slip detection and rapid grip control [3]. Such perception of micro-deformations is prevalent from initial contact [35] to active manipulation [2], and can apply to both tangential and rotational slip scenarios [36]. These findings highlight the importance of capturing subtle changes in contact mechanics for effective slip detection. Additionally, tribological studies have revealed specific characteristics of these phenomena. For example, the crawling motion of snails is accompanied by the transmission of stress waves [37]. Events of stress and strain drop have been detected during the slip of polymethyl methacrylate against glass under tangential loading [38]. Therefore, the transient changes in stress and strain and their distribution over the contact surface contain the spatio-temporal evolution characteristics of incipient slip.

Considering that accurate measurement of stress relies on known material parameters and contact characteristics, this article primarily focuses on using the spatio-temporal features of strain as the representation of incipient slip. Let the spatio-temporal evolution of strain on the contact surface be $f(x, t)$, where $x$ and $t$ denote the contact position and time, respectively. Discretizing the contact surface $S$ into $N$ surface elements, the strain at the $i$-th element $x_i$ is represented as $f_i(t) = f(x_i, t)$. Then, the local slip state at $x_i$ is

$$s_i(t) = g_{i,t}\left[\sum_{j=1}^{N}\sum_{\tau=0}^{t} f_j(t-\tau) \cdot w_{j,i}(\tau)\right] = \begin{cases} 1, & \text{slip} \\ 0, & \text{not slip} \end{cases}. \quad (1)$$

In Eq. (1), $w_{j,i}(\tau)$ denotes the influence function, which represents the weight of the stress-strain at micro-element $x_j$ on the influence of $x_i$ at moment $\tau$. $g(:)$ represents the nonlinear mapping relationship between the effect of $f(x, t)$ and the local slip state at $x_i$. In this article, for simplicity, we describe the local slip state as either 0 or 1, so $g(:)$ acts like a binary classifier. If the coefficients such as the friction coefficient on the contact surface are known, a continuous description of the local slip state can also be constructed using models like Coulomb friction or stochastic friction.

The time dependence of $w_{j,i}(\tau)$ arises from the propagation of tactile stimuli in space. Given that the impact typically dissipates within approximately 30 ms after contact [40], the effective influence occurs only within a limited time interval. For determining the local contact state, it is unnecessary to consider the entire incipient slip process but rather the variation in strain within a certain time interval $\Delta t$. Besides, slip effects are typically interpreted as the stress and strain release caused by localized shear failure processes [39], which means the slip state depends only on the stress-strain state at the current position. Thus, two assumptions are introduced:
1) The local slip state depends only on the strain state within a finite period.
2) The local slip state depends only on the local strain state within the neighborhood.

Based on the above assumptions, Eq. (1) can be simplified as:

$$s_i(t) = g_{i,t}\left[\sum_{\Delta t} f_i(t-\tau) \cdot w_{i,i}(\tau)\right] =$$
$$g_{i,t}^*\{[f_i(t), f_i(t-1), \dots, f_i(t-\Delta t)]^T\}. \quad (2)$$

Eq. (2) reflects the temporal characteristics of incipient slip. Here, $g_{i,t}^*(:)$ represents the mapping relationship between the state changes of $f_i(t)$ and $s_i(t)$ within the time interval $\Delta t$. The perception of the temporal characteristics of incipient slip depends on establishing the potential relationship between the transient changes in strain and the switching of local slip states, while excluding the influence of friction and geometric characteristics. Additionally, the form of $g_{i,t}^*(:)$ varies under different loading and motion conditions. Considering the temporal characteristics of incipient slip allows differentiation of different contact states, improving the adaptability of the detection method to various conditions.

To investigate the extent of incipient slip across the entire surface, we define it as the low-dimensional representation of the high-dimensional information of incipient slip:

$$\psi(t) = \eta_t\{[s_1, s_2, \dots, s_N]^T\}. \quad (3)$$

The selection of $\psi$ depends on the requirements of task and the characteristics of the working conditions. In existing studies, features such as the contact force ratio [20], safety margin [25], and stick-slip ratio [27] have been used. Besides, $\eta_t(:)$ represents the mapping relationship between the local slip states at each position on the contact surface at time $t$ and the overall degree of incipient slip on the contact surface. At a given moment, a complete representation of incipient slip requires considering the local slip states at all positions on the contact surface. Thus, Eq. (2) reflects the spatial characteristics of incipient slip. Due to the geometric and frictional differences in the contact between objects of different shapes, materials, and degrees of hardness and softness, the spatial distribution of the localized slip states may differ even for the same loading process. As a result, detecting the spatial characteristics of incipient slip ensures high adaptability of the method to various objects.

Based on the above discussion, the definition of the spatio-temporal characteristics of incipient slip is given as follows:
1) *Temporal characteristics*: The relationship between the strain state at each location on the contact surface and the local slip state.
2) *Spatial characteristics*: The distribution of local slip states on the contact surface and their relationship with the degree of incipient slip.

The subsequent parts of this section further discuss the theoretical development and measurement methods for the spatio-temporal characteristics during incipient slip.

### B. Spatial Characteristics of Slip Field

In this subsection, we analyze the evolution of the spatio-temporal characteristics of the slip field during incipient slip and examine its relationship with the strain field. Consider the contact model shown in Fig. 2(a). Under the action of normal force $F_N$, a spherical elastic body $s$ (tactile fingertip) is compressed along the -z direction and contacts a rigid plane $b$ (contact object). The tangential force $F_T$ is applied along the x-direction, causing elastic deformation on the contact surface. According to the derivation in Appendix A, the slip field can be expressed as:

$$s(r) = s_x =$$
$$\begin{cases} H_1 \cdot \left\{ \begin{array}{l} \left[1 - \dfrac{2}{\pi} sin^{-1}\left(\dfrac{r_c}{r}\right)\right] \cdot \left(1 - \dfrac{2r_c{}^2}{r^2}\right) \\ + \dfrac{2}{\pi}\dfrac{r_c}{r}\left(1 - \dfrac{r_c{}^2}{r^2}\right)^{1/2} \end{array} \right\}, \ r \in [0, r_c], \\ 0, r \in (r_c, r_a) \end{cases} \quad (4)$$

$$where \quad H_1 = \frac{3(2-\nu)\mu F_N}{16 G r_a}. \quad (5)$$

Here, $r_a$ and $r_c$ represent the radii of the contact region and the stick region, respectively. $\mu$ is the friction coefficient, $G$ denotes the shear modulus, and $\nu$ represents the Poisson's ratio. According to [31], if the deformation of object $b$ is also considered, a similar relation (with different coefficients) can also be obtained. Ignoring discontinuities in the function, the derivative of (4) is obtained as:

$$\frac{ds(r)}{dr} = \begin{cases} H_2 \cdot 4\dfrac{r_c{}^3}{r^3}\left(\dfrac{\pi}{2} - sin^{-1}\left(\dfrac{r_c}{r}\right)\right), \ r \in [0, r_c], \\ 0, r \in (r_c, r_a) \end{cases} \quad (6)$$

$$where \quad H_2 = \frac{2H_1}{\pi r_c} = \frac{3(2-\nu)\mu F_N}{8\pi G r_a r_c}. \quad (7)$$

We define the dimensionless functions $\varphi_1(\beta)$ and $\varphi_2(\beta)$:

$$s = H_1 \varphi_1(\beta), \quad (8)$$

$$ds/dr = H_2 \varphi_2(\beta), \quad (9)$$

where

$$\varphi_1(\beta) = \begin{cases} \left[1 - \dfrac{2}{\pi}sin^{-1}(\beta^{-1})\right] \cdot (1 - 2\beta^{-1}) + \\ \dfrac{2}{\pi}\beta^{-1}(1 - \beta^{-2})^{1/2}, \ \beta \in [0,1] \\ 0, \ \beta \in (1, \alpha) \end{cases}, \quad (10)$$

$$\varphi_2(\beta) = \begin{cases} \dfrac{4}{\beta^3} \cdot \left(\dfrac{\pi}{2} - sin^{-1}(\beta^{-1})\right), \ \beta \in [0,1) \\ 0, \ \beta \in (1, \alpha) \end{cases}, \quad (11)$$

$$where \quad \beta = r/r_c, \alpha = r_c/r_a. \quad (12)$$

$\varphi_1(\beta)$ and $\varphi_2(\beta)$ represent the dimensionless distributions of slip and its spatial derivative, respectively, as shown in Fig. 2(b). In the stick region, the slippage is always zero. Within the slip region, the slip magnitude increases with distance from the stick-slip boundary, while the spatial derivative first increases and then decreases. Notably, near the stick-slip boundary ($\beta = 1.17$), the spatial derivative reaches an extreme value. Moving outward from the contact center, the spatial derivative remains nearly zero within the stick region. However, as it approaches the stick-slip boundary, it shows a marked increase, rapidly reaching its maximum before rapidly decreasing as the distance from the contact center continues to grow. Therefore, the extreme distribution of the spatial derivative approximately reflects the outer envelope of the stick region.

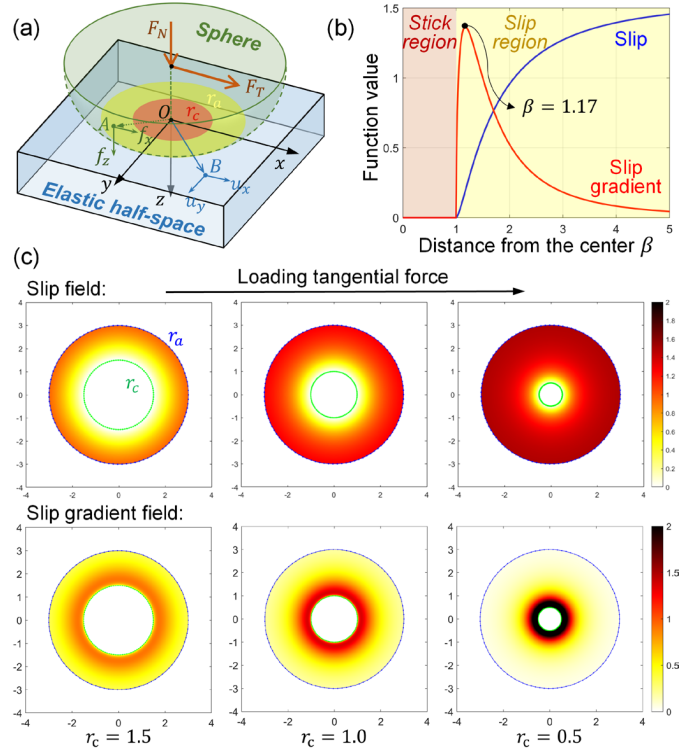

**Fig. 2.** (a) Contact model for incipient slip [31]. (b) Relationship functions of slip and its spatial derivative with the distance from the contact center. (c) Evolution of the slip field and its spatial derivative distribution as the tangential force increases. Here, $r_a$ (blue dashed line) represents the contact region boundary, and $r_c$ (green solid line) indicates the stick-slip boundary.

Consider the following contact process: keeping the normal contact force $F_N$ constant, gradually increasing $F_T$ from zero. Set $r_a = 3$, and observe the evolution of the slip field and its spatial derivative as the stick radius $r_c$ decreases from 1.5 to 0.5. The distributions of $\varphi_1(\beta)$ and $\varphi_2(\beta)$ during this process are shown in Fig. 2(c). As the tangential load increases, the stick region shrinks, and the slip field progressively spreads from the outer edge inward. The relationship between the slip distribution and the stick-slip boundary is not intuitive, as less relative motion accumulates closer to the stick region. In contrast, the spatial derivative of the slip field propagates inward in a wave-like pattern, with the wavefront accurately reflecting the stick-slip boundary. The location of the extreme value of the spatial derivative corresponds to the newly developed slip region. Therefore, the spatial derivative of slip effectively characterizes the spatial features of incipient slip.

The spatial derivative of slip, in terms of its physical significance, is equivalent to the difference in the plane strain field on the contact surface. Considering that both the contact bodies $s$ and $b$ are elastic bodies, the spatial derivative of slip can be expressed as:

$$\frac{\partial s}{\partial n} = \frac{\partial u_n^b}{\partial n} - \frac{\partial u_n^s}{\partial n} = \varepsilon_n^b - \varepsilon_n^s, \quad n \in \{x, y\}, \tag{13}$$

where $\varepsilon_n^b$ and $\varepsilon_n^s$ represent the components of the plane strain in the $n$-direction on contact bodies $s$ and $b$, respectively. According to [27], for soft contact，$\varepsilon_n^b$ and $\varepsilon_n^s$ satisfy the relation:

$$\varepsilon_n^b = -\frac{G^s}{G^b}\varepsilon_n^s + l_{nn}, \tag{14}$$

$$\text{where} \quad l_{nn} \equiv const. \tag{15}$$

Substituting Eq. (14) and Eq. (15) into Eq. (13) yields

$$\frac{\partial s}{\partial n} = -\frac{G^s + G^b}{G^b}\varepsilon_n^s + l_{nn}. \tag{16}$$

Eq. (16) indicates that even when considering finite soft contact, the linear positive correlation between the plane strain and the spatial derivative of slip still holds. This suggests that the plane strain field on the contact surface approximately represents the spatial derivative of slip, and its evolution exhibits wave-like propagation from the outside inward. Thus, the wavefront of the strain rate can be used to reflect the newly developed incipient slip region.

## C. Temporal Characteristics of Strain Rate

This subsection further analyzes the relationship between the plane strain rate and incipient slip, and develops a quantification method for the time characteristics of early slip. Previous studies proposed that a soft fingertip model composed of virtual elastic beams can be used to study various deformations of the fingertip during sliding motion [41], as shown in Fig. 3(a). In this model, by observing the dynamic motion of free ends of the beams constrained by Coulomb's friction law, the transient response during the fingertip's incipient slip process can be evaluated. For the sake of simplicity, we simplify the damping term in the Voigt model (i.e., neglecting viscoelasticity) and model the interaction between beams as linear elastic elements (springs) with stiffness $k$.

The normal force $F_N$ and tangential force $F_T$ are applied to the object, causing it to move relative to the contact surface of the fingertip at a constant velocity $\dot{u}$. $u$ represents the overall displacement of the fingertip. Consider the contact between the $i$-th beam and the object, as well as the interaction with the ($i$-1)-th and ($i$+1)-th adjacent beams. Let the generalized displacements of the three free ends relative to the fingertip pad be $u_{i-1}$, $u_i$, and $u_{i+1}$, respectively. The interaction forces between the linear elastic elements of adjacent beams acting

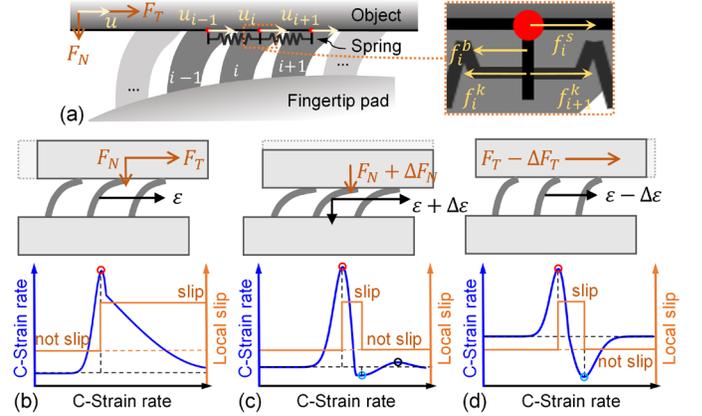

**Fig. 3.** (a) Soft fingertip model based on a virtual elastic beam. (b)-(c) Strain rate extreme events. The first row shows a schematic of the contact state, and the second row illustrates the temporal variation of the plane strain rate during the loading process. (b) Case 1: unidirectional tangential contact with constant normal force; (c) Case 2: increasing normal force based on Case 1; (d) Case 3: reducing tangential force based on Case 1.

on the $i$-th beam can be calculated as:

$$f_{i+1}^k = k(u_{i+1} - u_i), \tag{17}$$

$$f_i^k = k(u_i - u_{i-1}). \tag{18}$$

In addition, the tangential force due to bending of the beam is expressed as

$$f_i^b = b_i u_i, \tag{19}$$

where $b_i$ denotes the bending stiffness of the $i$-th beam. Suppose the interface frictional force is $f_i^s$. Under the quasistatic assumption, the force balance equation for the $i$-th beam can be expressed as:

$$f_i^k - f_{i+1}^k + f_i^b = f_i^s \leq f_{i,m}^s. \tag{20}$$

Let the spacing between the beams be $\Delta$. For the one-dimensional model, the plane strain $\varepsilon$ is the gradient of the deformation field. Therefore, Eq. (20) can be expressed as:

$$k\Delta \cdot (\varepsilon_i - \varepsilon_{i+1}) + b_i u_i = f_i^s. \tag{21}$$

Let $\delta$ be the plane strain rate. Then, taking the derivative of Eq. (21) with respect to time gives:

$$k\Delta \cdot (\delta_i - \delta_{i+1}) + b_i \dot{u}_i = \dot{f}_i^s. \tag{22}$$

At the beginning, the ($i$-1)-th, $i$-th, and ($i$+1)-th beams are all in the sticking state. Therefore, $\delta_i$ and $\delta_{i+1}$ are both zero, and $\dot{f}_i^s$ increases in a constant proportion of $b_i \dot{u}$. As the tangential force increases, the nodes closer to the outer side of the contact region gradually reach the Coulomb friction limit. Without loss of generality, assume that local slip occurs first in the region to the left of the ($i$-1)-th beam. Consider the change in state at the $i$-th beam from sticking to local slip. Since the ($i$+1)-th beam remains in the stick state (i.e., $\dot{u}_{i+1} =$ $\dot{u}$ ) throughout this process, we have $\delta_{i+1} = \dot{u} - \dot{u}_i$. Substituting this into Eq. (22) gives:

$$\delta_i = \frac{\dot{f}_i^s}{k\Delta} + \dot{u} - \frac{k\Delta + b_i}{k\Delta}\dot{u}_i. \tag{23}$$

1) Before the $i$-th beam slips, $\dot{u}_i = \dot{u}_{i+1} = \dot{u}$ holds. Thus,

$$\delta_i = \frac{\dot{f}_i^s}{k\Delta} - \frac{b_i}{k\Delta}\dot{u}. \tag{24}$$

At the beginning, $\dot{f}_i^s = b_i\dot{u}$, so $\delta_i = 0$ still holds. According to the feature of localized displacement phenomenon, $\dot{u}_{i-1}$ should gradually decrease from $\dot{u}$ until it equals zero. Therefore, $\delta_i$ will gradually increase from zero. This process is accompanied by an increase in $\dot{f}_i^s$, until $\dot{f}_i^s$ reaches $f_{i,m}^s$. Since the rate of increase in $f_i^s$ becomes larger the closer it gets to the Coulomb friction limit, the $i$-th beam will quickly reach the local slip state.

2) After the $i$-th beam begin slipping, $\dot{f}_i^s$ quickly decreases to zero. Assuming that the $(i$-1)-th beam has already fully slipped, i.e., $\dot{u}_{i-1} = 0$, we have

$$\delta_i = \frac{\dot{f}_i^s}{2k\Delta + b_i} + \frac{k\Delta}{2k\Delta + b_i}\dot{u}. \tag{25}$$

Therefore, $\delta_i$ will quickly decrease until $\delta_{th} = \frac{k\Delta}{2k\Delta + b_i}\dot{u}$.

3) When the $i$-th beam has fully slipped and the $(i$-1)-th beam begins to slip, $\dot{f}_i^s = 0$. Thus,

$$\delta_i = \frac{k\Delta}{2k\Delta + b_i}\dot{u}_{i+1}. \tag{26}$$

At this point, $\delta_i$ will decrease at a slower rate until it eventually equals zero.

Therefore, based on the localized displacement phenomenon in the slip process, the extreme events of the plane strain rate in the positive semi-axis correspond to the occurrence of localized slip. This process describes Case 1, as shown in Fig. 3(b). For unidirectional relative tangential contact, the plane strain rate at a point on the contact surface first increases and then decreases as the tangential load increases, reaching a maximum value when localized slip occurs. Additionally, we discuss the strain-rate extreme events of two other typical working conditions:

1) Case 2: After tangential loading, an increase in normal force is applied [see Fig. 3(c)]. When tangential loading stops increasing, the velocity of the free end of the beam reverses because the direction of the frictional force should remain unchanged to counterbalance the bending force of the beam. According to Eq. (26), the plane strain rate will decrease to below zero. When the contact state switches, the $i$-th beam transitions from slip state to stick state and experiences compression. For the virtual elastic beam model used, normal compression also leads to an increase in tangential displacement. Therefore, this process is similar to the tangential loading scenario described by Eqs. (25) and (26). As a result, the plane strain rate will reach a secondary peak lower than the main peak.

2) Case 3: After tangential loading, decrease the tangential force to zero [see Fig. 3(d)]. Due to the reversal of $\dot{u}$, the plane strain rate will undergo the reverse process of positive loading. In the negative space, $\delta_i$ first decreases and then increases, until it returns to zero. The minimum value occurs when the contact state switches (i.e., from slip to stick). In addition, when the tangential force is reduced to the reverse direction, the strain rate will undergo a new process of Case 1 and reach the local slip for the second time.

As a result, the strain-rate extreme events in the negative semi-axis correspond to the end of localized slip. After slip occurs, whether the change in contact state is caused by an increase in normal force or a decrease in tangential force, the minimum value of the plane strain rate corresponds to the end of localized slip. If no transition from stick to slip occurs, the minimum event will not be observed in the negative space, as shown in Fig. 3(b).

In summary, the extreme events of strain rate reflect the temporal characteristics of incipient slip. Specifically, the positive extreme represents the transition from localized stick to slip, while the negative extreme represents the transition from localized slip to stick. Combined with the derivation in Section II-B, the propagation of strain rate extreme events can be used as indicators to characterize the spatial features during slip. In practical applications, by detecting the strain rate at different positions on the contact surface of the robotic skin through tactile sensors, the spatio-temporal distribution of incipient slip can be obtained.

### D. Detection of Strain-Rate Extreme Events

The sensing of strain rate requires measurement equipment capable of detecting contact deformation. Suppose there are $n \times m$ effective measurement points on the contact surface. At a certain moment, the coordinate of the $(i, j)$-th measurement point $p_{i,j} = \mathbf{r}(i, j) = (x, y, z)$ , and its displacement is $\mathbf{u}(i, j) = (u_x, u_y, u_z)$. Thus, the deformation field gradient at $p_{i,j}$ can be estimated as:

$$\begin{cases} \dfrac{\partial u_k}{\partial x} = \dfrac{u_k(i+1, j) - u_k(i-1, j)}{x(i+1, j) - x(i-1, j)} \\ \dfrac{\partial u_k}{\partial y} = \dfrac{u_k(i, j+1) - u_k(i, j)}{y(i, j+1) - y(i, j+1)} \\ \qquad \text{where } k \in \{x, y\}. \end{cases} \tag{27}$$

Based on Eq. (27), the strain rate in different directions can be determined. Moreover, to apply the strain-rate metric in practice, three issues need to be addressed:

1) Finite deformation condition: The fingertip undergoes overall motion and rotation, which goes beyond the small deformation assumption. Similar to [33]-[35], the strain should be expressed in the Green-Lagrange form:

$$\varepsilon_{xx} = \frac{\partial u_x}{\partial x} + \frac{1}{2}\left[\left(\frac{\partial u_x}{\partial x}\right)^2 + \left(\frac{\partial u_y}{\partial x}\right)^2\right], \tag{28}$$

$$\varepsilon_{yy} = \frac{\partial u_y}{\partial y} + \frac{1}{2}\left[\left(\frac{\partial u_x}{\partial y}\right)^2 + \left(\frac{\partial u_y}{\partial y}\right)^2\right], \tag{29}$$

$$\varepsilon_{xy} = \frac{1}{2}\left(\frac{\partial u_x}{\partial y} + \frac{\partial u_y}{\partial x}\right) + \frac{1}{2}\left[\frac{\partial u_x}{\partial x}\frac{\partial u_x}{\partial y} + \frac{\partial u_y}{\partial x}\frac{\partial u_y}{\partial y}\right], \quad (30)$$

where $\varepsilon_{xx}$ and $\varepsilon_{yy}$ are the axial strain components aligned with the x-axis and y-axis, respectively, and $\varepsilon_{xy}$ denotes the shear strain.

*2) Coordinate system consistency:* Since the direction of the tangential load is unknown, the defined plane strain should be independent of the coordinate system. Let the strain matrix be represented in principal strain form through eigenvalue decomposition:

$$\boldsymbol{\varepsilon} = \begin{bmatrix} \varepsilon_{xx} & \varepsilon_{xy} \\ \varepsilon_{xy} & \varepsilon_{yy} \end{bmatrix} = \begin{bmatrix} \beta_{11} & \beta_{12} \\ \beta_{21} & \beta_{22} \end{bmatrix}\begin{bmatrix} \varepsilon_1 & 0 \\ 0 & \varepsilon_2 \end{bmatrix}\begin{bmatrix} \beta_{11} & \beta_{12} \\ \beta_{21} & \beta_{22} \end{bmatrix}^{-1}, (31)$$

The principal strain decomposition is equivalent to a rotation of the reference frame, ensuring coordinate system consistency while canceling out the shear strain. The obtained principal strains, $\varepsilon_1$ and $\varepsilon_2$, correspond to the maximum tensile strain and compressive strain, respectively.

*3) Elastic contact (non-planar geometry):* The above derivations assume that the contact surface is flat. In reality, for common contacts between soft fingertips and deformable objects, the contact surface is not flat and should instead be represented by a shape function $\boldsymbol{S}(x, y, z)$. For each position $\boldsymbol{p} = p(x, y, z)$, the normalized normal direction of $\boldsymbol{S}$ at $\boldsymbol{p}$ is

$$\boldsymbol{n} = [n_x, n_y, n_z]^T = \frac{1}{\|\nabla \boldsymbol{S}\|} \cdot [\frac{\partial \boldsymbol{S}}{\partial x}, \frac{\partial \boldsymbol{S}}{\partial y}, \frac{\partial \boldsymbol{S}}{\partial z}]^T, \quad (32)$$

In the neighborhood of $\boldsymbol{p}$, the contact patch can be approximated as a plane with a normal direction $\boldsymbol{n}$. For the sake of simplicity, assume that the strain components parallel to this plane dominate. Therefore, the principal strains in the neighborhood can be approximated as:

$$\begin{bmatrix} \varepsilon_1^* \\ \varepsilon_2^* \end{bmatrix} = \frac{1}{\hat{\boldsymbol{z}} \cdot \boldsymbol{n}} \cdot \begin{bmatrix} \varepsilon_1 \\ \varepsilon_2 \end{bmatrix} = \frac{\|\nabla \boldsymbol{S}\|}{\partial \boldsymbol{S}/\partial z} \cdot \begin{bmatrix} \varepsilon_1 \\ \varepsilon_2 \end{bmatrix}. \quad (33)$$

For a continuous sequence of tactile information, the principal strain rate is represented as the principal strain increment between two adjacent time frames:

$$\begin{bmatrix} e_1(t) \\ e_2(t) \end{bmatrix} = \frac{1}{\Delta t}\begin{bmatrix} \varepsilon_1^*(t + \Delta t) - \varepsilon_1^*(t) \\ \varepsilon_2^*(t + \Delta t) - \varepsilon_2^*(t) \end{bmatrix}, \quad (34)$$

where $\Delta t$ represents the time interval. To quantify the Euclidean distance between the principal strain rate vectors, the L2-norm is chosen to define the plane characteristic strain rate (abbreviated as C-Strain rate):

$$\delta = \sqrt{e_1^2 + e_2^2}. \quad (35)$$

The characteristic strain (abbreviated as C-Strain) is featured by describing the overall properties of the plane strain. Its value is independent of the orientation of the coordinate axes, making it a unified measure of the evolution of incipient slip under different directions of tangential loads and torques. In the subsequent discussions, unless specifically mentioned, we default to using the C-Strain rate as a measure of the spatio-temporal characteristics of incipient slip.

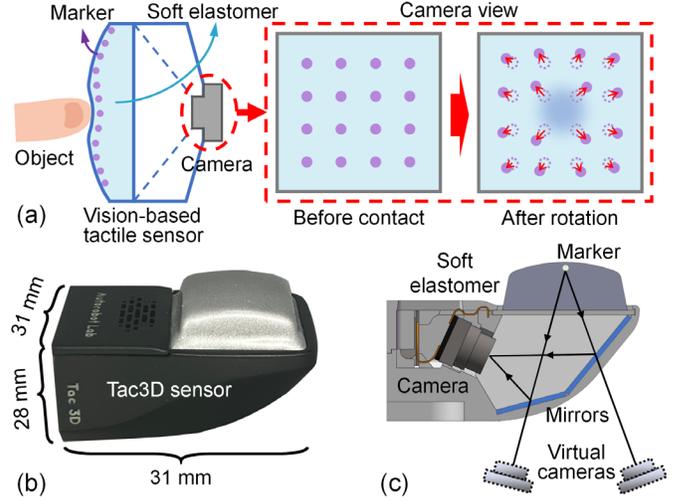

**Fig. 4.** Vision-based sensing method. (a) Contact deformation detection based on marker displacement method. (b) Tac3D tactile sensor. (c) Principle of virtual binocular vision.

Furthermore, based on real-time peak detection algorithms (such as the z-score based peak detection method [42] used in this article), the strain rate extreme events discussed in Section II -C (i.e., the local maximum and minimum peaks of the real-time signal) can be identified. This allows for the acquisition of the response distribution signals of localized slip initiation and termination on the contact surface. In addition, note that extreme values in the positive direction need to be filtered by setting a threshold to exclude maximum events unrelated to slip, which are caused by normal contact (see the sub-peak in Fig. 3(c)). In this article, 0.1 times the ratio of the current strain to the measurement frame rate (i.e., sampling period) is chosen as the threshold.

Note that the proposed method does not limit the types of tactile sensors. Any device capable of measuring contact deformation can be applied to the above detection process. In the simulation experiments (Sections III), we directly obtained the deformation field and contact geometry on the contact surface, and calculated the C-Strain in accordance with Eqs. (27)-(35). In the practical experiment (Sections IV and V), we chose the Tac3D tactile sensor as the implementation equipment. The sensing principle based on marker displacement of Tac3D is shown in Fig. 4(a). By tracking the movement of marker points, the contact deformation could be discretely sampled [22]. Fig. 4(b) shows the structure of the Tac3D sensor. It used a virtual binocular vision system that employs a single camera and a mirror to simulate the effect of binocular cameras, enabling 3-D measurement in a compact space. It is an efficient way to obtain the contact geometric information. Ultimately, the original deformation field measured by the Tac3D sensor is calculated as the C-Stran ratio. The onset and spatial propagation of localized slip were determined based on the extreme events of the C-Strain, which led to the determination of the spatio-temporal properties of incipient slip across the contact surface.

## III. Simulation

### A. Finite Element Simulation Model

In this subsection, we used finite element analysis (FEA) simulations in Abaqus to validate the proposed detection method and compared it with existing methods. Fig. 5(a) shows the constructed simulation model to simulate the contact scenario of a parallel fingertip grasping an object. The fingertip was modeled as a cylinder with a radius of 15 mm and a thickness of 5 mm, and its Young's modulus was set as 1 MPa. The objects can be replaced with different shapes and degrees of stiffness. In the simulation, the fingertip was in contact with the object under normal force $F_N$, tangential force $F_T$, and torque $M$. The relative friction coefficient (Coulomb friction) was set to $\mu$. Due to the symmetrical contact state, only the incipient slip evolution on one fingertip needs to be considered.

In the first part of the experiment (Sections III-B and III-C), a rigid cuboid object (40 mm × 40 mm × 20 mm) was considered. The constructed contact method is shown in Fig. 5(b). First, a normal force $F_N$ was applied to provide preload. At this point, no slip was analyzed due to the absence of tangential action. Then, a tangential force $F_T$ or a torque $M$ was applied. Four different conditions, as shown in Table I, were selected for analysis with the loading processes illustrated in Fig. 5(c). By controlling variables, the effects of different conditions on incipient slip detection were compared: the difference between Condition 1 and Condition 2 is the friction coefficient, Condition 1 and Condition 3 were used to compare the normal force, and Condition 4 demonstrates the applicability of the proposed method to torsional conditions. During the tangential contact process, the normal force remains constant. By recording the contact deformation and distributed force at each analysis step, the results of different slip detection methods and the ground truth of incipient slip were then calculated for comparison.

### B. Strain Evolution during Incipient Slip

1) *Strain field components*: Taking Condition 1 as an example, the heatmaps in Fig. 6(a)-(c) show the evolution of the strain field within the contact area in three directions. The simulation process illustrates the progression from the initial state of tangential loading to the point where the strain rate returns to nearly zero. The results indicate that a gradual strain wave propagates from the periphery to the center of the contact area. For the strain $\varepsilon_{xx}$ in the x-direction, compressive strain and tensile strain dominate at opposite ends of the stick region. Specifically, compressive waves occur at the end closer to the shear direction, while tensile waves appear at the other end. In contrast, the contributions of $\varepsilon_{yy}$ and $\varepsilon_{xy}$ are relatively minor. $\varepsilon_{xy}$ accumulates laterally along the shear direction, while $\varepsilon_{yy}$ shows significant changes only in the compressed part of the stick region.

Notably, the spatial distribution characteristics of the strain are consistent with physiological phenomena observed in human finger pads [33]-[36]. This issue reflects the agreement between simulation and reality. As the stick region shrinks, the

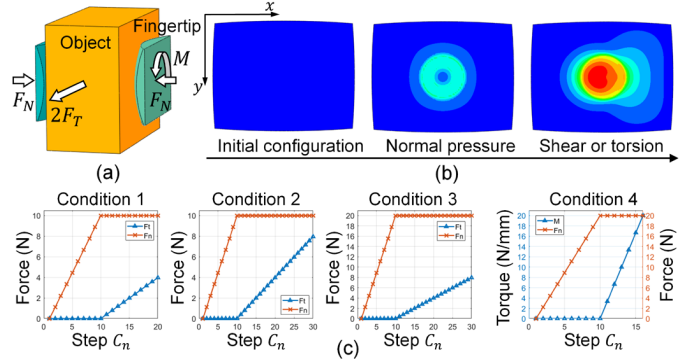

**Fig. 5.** (a) Simulation environment and model. (b) Contact process: a normal force was applied first, followed by a tangential force or torque. The color distribution in the figure represents the von-Mises stress. (c) Variation of the applied forces with analysis steps under four different contact conditions.

### TABLE I
### FOUR TYPES OF CONTACT CONDITIONS

| Conditions | $\mu$ | $F_N/N$ | $2F_T/N$ | $M/N \cdot mm^{-1}$ |
|---|---|---|---|---|
| **1** | 0.4 | 10 | 7.95 | 0 |
| **2** | 0.8 | 10 | 15.9 | 0 |
| **3** | 0.4 | 20 | 15.9 | 0 |
| **4** | 0.4 | 20 | 0 | 20 |

strains in different directions gradually propagate inward. Also, the propagation fronts reflect the contributions of slip and tangential loading in the corresponding components. These results also suggest that relying solely on a single strain component cannot fully capture the spatial characteristics of incipient slip. In contrast, using the C-Strain instead of principal strain under the global coordinate system helps to address this limitation.

2) *C-Strain rate field*: Figs. 6(d)-(g) illustrate the evolution of the C-Strain rate under four different conditions. The heatmaps in each column correspond to the results of the analysis steps where the ratio of concentrated forces, $F_T/\mu F_N$, is equal across the four conditions. This ratio describes the proximity to macro slip on the contact surface and serves as a fundamental stability indicator. Therefore, the spatial distribution of the C-Strain rate can be directly compared under the same stability level across different conditions. The results show that, under different contact conditions, the distribution of the C-Strain rate evolves similarly with the overall slip degree on the contact surface: the strain rate waves gradually converge from the periphery to the center, and the absolute value progressively increases.

Besides, comparing Conditions 1-3 reveals that the primary differences are in the absolute values of the C-Strain rate, which are influenced by the friction coefficient and normal force. Specifically, a larger friction coefficient results in a higher C-Strain rate at the same stability level, while a larger normal force slightly reduces the C-Strain rate, although the difference is insignificant. In others word, the differences in the evolution of C-Strain rate reflect the effects of varying friction and contact conditions. This observation aligns with the characteristics of human tactile perception and grasp force control under different friction coefficients and gripping forces [2]-[3]. Additionally, the results of Condition 4 indicate that the evolution of C-Strain rate caused by torsional loading is similar to that under tangential loading. Since the C-Strain rate eliminates directional attributes through consistent changes, it responds equally to slip caused by either translational or rotational loading.

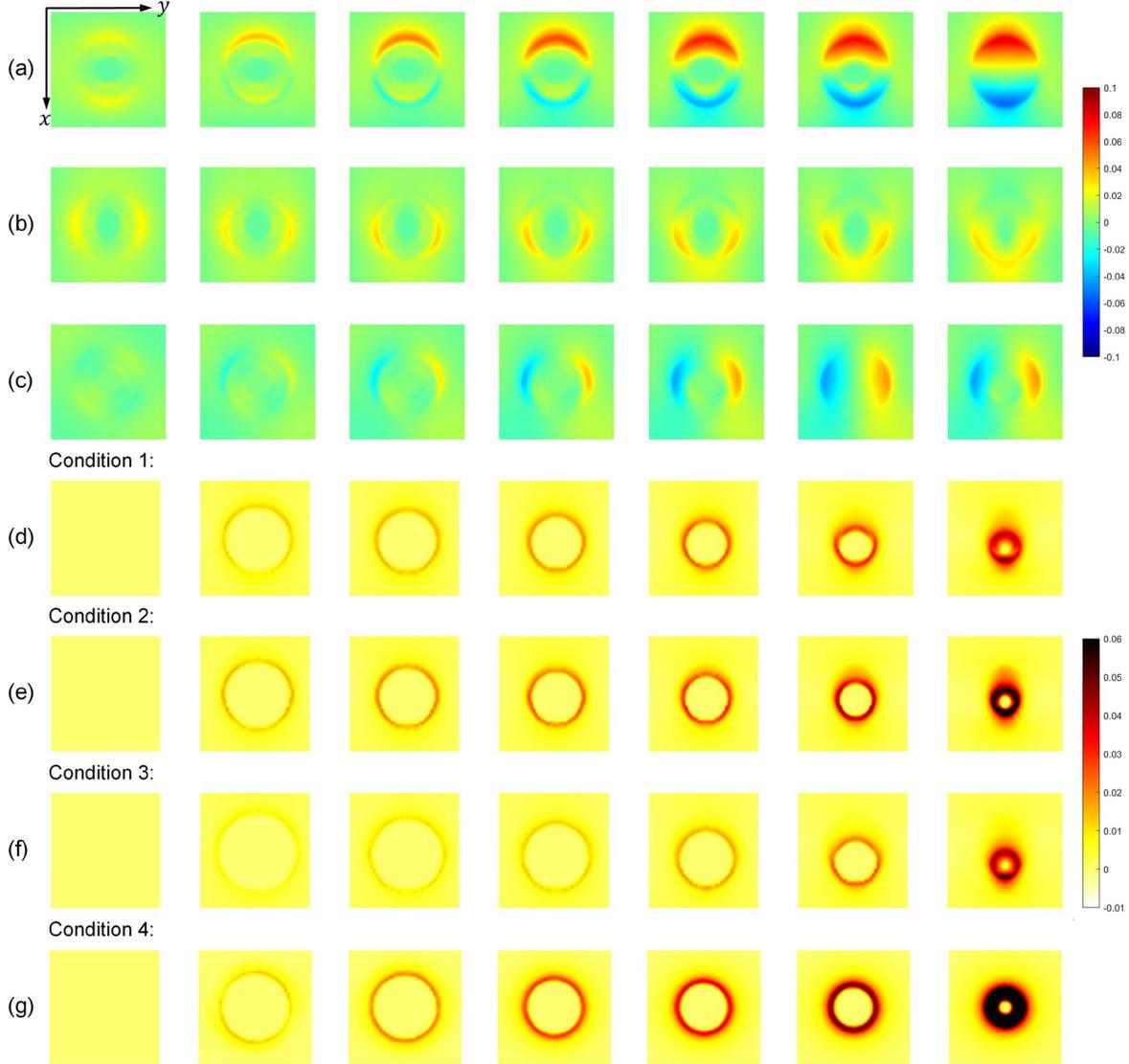

**Fig. 6.** Evolution of spatial distribution in Condition 1. (a)-(c) Evolution of strain field components. (a): Normal strain $\varepsilon_{xx}$. (b): Normal strain $\varepsilon_{yy}$. (c): Shear strain $\varepsilon_{xy}$. (d)-(g) Evolution of C-Strain rate fields under four contact conditions.

### C. Effectiveness Evaluation

1) *Temporal characteristics*: As previously demonstrated, C-Strain rate extreme events serve as responses to the temporal characteristics of incipient slip. To evaluate the effectiveness of the proposed method, we selected three reference points $p_a$, $p_b$, and $p_c$, as shown in Fig. 7(a), to analyze the time evolution characteristics of the C-Strain rate at each point. Similar to the concentrated force ratio used earlier, the ground truth of incipient slip was calculated based on the contact factor defined in [16]:

$$cf = f_t / \mu f_n, \qquad (36)$$

where $f_t$ and $f_n$ represent the micro-element tangential force and normal force at the given location, respectively. When the contact coefficient $cf$ is less than 1, the micro element at this position is in the stick state under Coulomb friction constraints; otherwise, local slip occurs. Additionally, a threshold of 0.001 was set to reduce the impact of simulation errors. Figs. 7(b)-(c) show the measurement results under three conditions. Each set of results includes the contact coefficient, the C-Strain rate, and the local slip state change

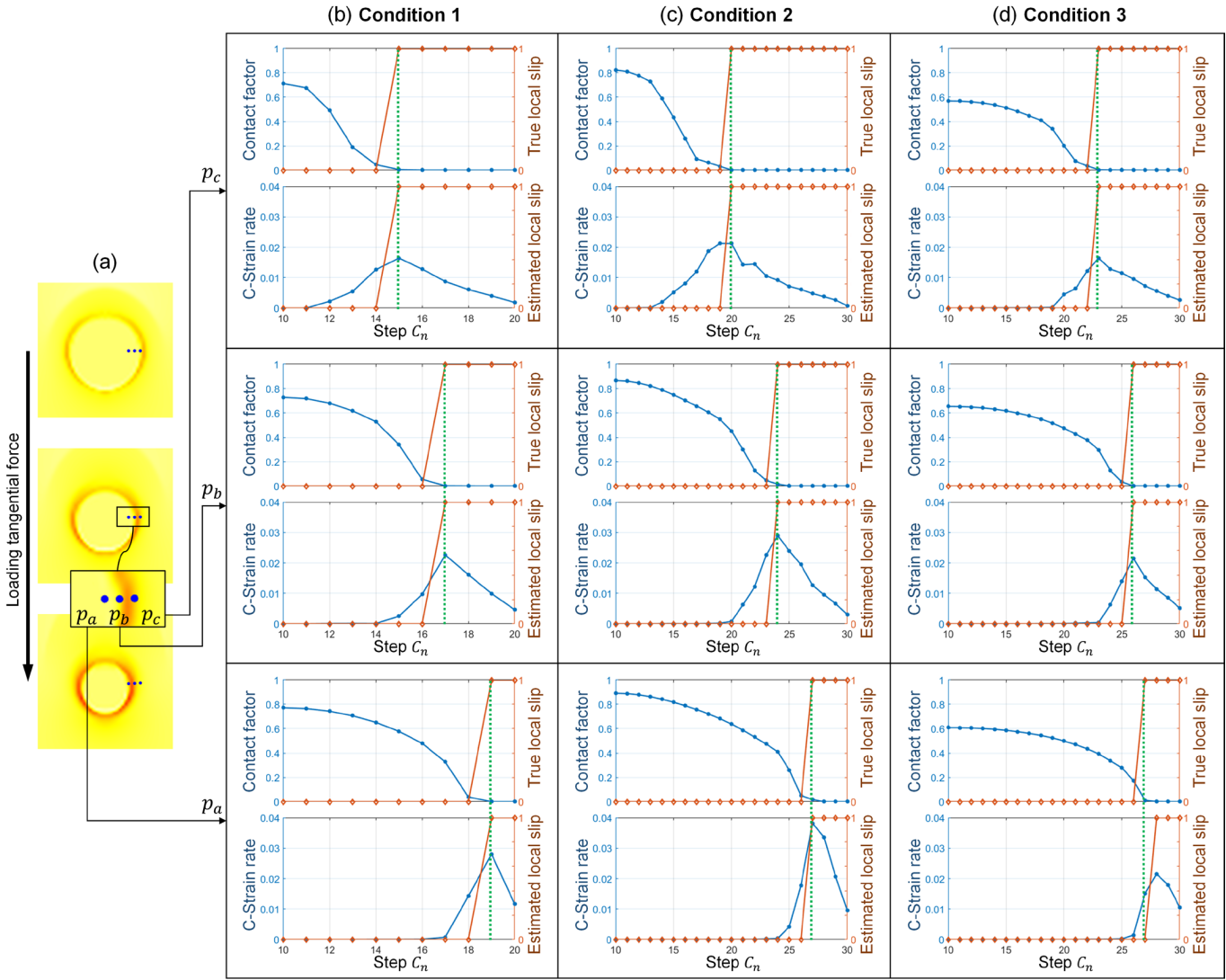

**Fig. 7.** Evaluation of incipient slip detection in the temporal dimension. (a) Selection of reference points. (b)-(c) The relationship between C-Strain rate extreme events and local slip state changes under three different conditions. For each set of subplots, the upper part shows the variation of the contact coefficient and the actual local slip state over the analysis steps, while the lower part displays the C-Strain rate and the estimated local slip state. The green dashed lines indicate the consistency between the estimated and actual values.

calculated based on these two time series signals.

The results indicate that, except for the estimated slip at position $p_a$ in Condition 3, which lags behind the actual slip by one analysis step, the rest match well (see the green line in the figures). Therefore, the temporal characteristics of the C-Strain rate extreme events can effectively describe the occurrence of local slip. Since this metric depends solely on local strain and does not require other measurement points, it can avoid the need for the sticking center region or frictional coefficient as required by existing methods [26], [27]. Furthermore, comparing the results under three conditions, it is found that the pulse peak of the C-Strain rate is related to the friction coefficient, while the phase is less influenced by it. Specifically, the rougher the contact surface, the greater the response of the C-Strain rate. Although the pulse peak is less related to normal contact, its phase is correlated with the normal force: the larger the normal force, the more delayed the occurrence of local slip at the same position. Different contact factors jointly influence the temporal characteristics of the C-Strain rate.

2) *Spatial characteristics*: The effectiveness of the method in the spatial dimension was further evaluated, as shown in Fig. 8. Fig. 8(a) demonstrates the real stick-slip region and the segmented stick-slip region based on the proposed method under four different conditions. For various conditions, the real stick-slip regions align well with the estimated ones. As shown in Fig. 6(d)-(g), the wavefront of the evolution of the C-Strain rate represents the boundary between the stick region and the slip region, and the region with significant C-Strain rate is approximately the newly formed slip region. This is primarily because the simulation considers a rigid contact object, resulting in the C-Strain rate within the stick region

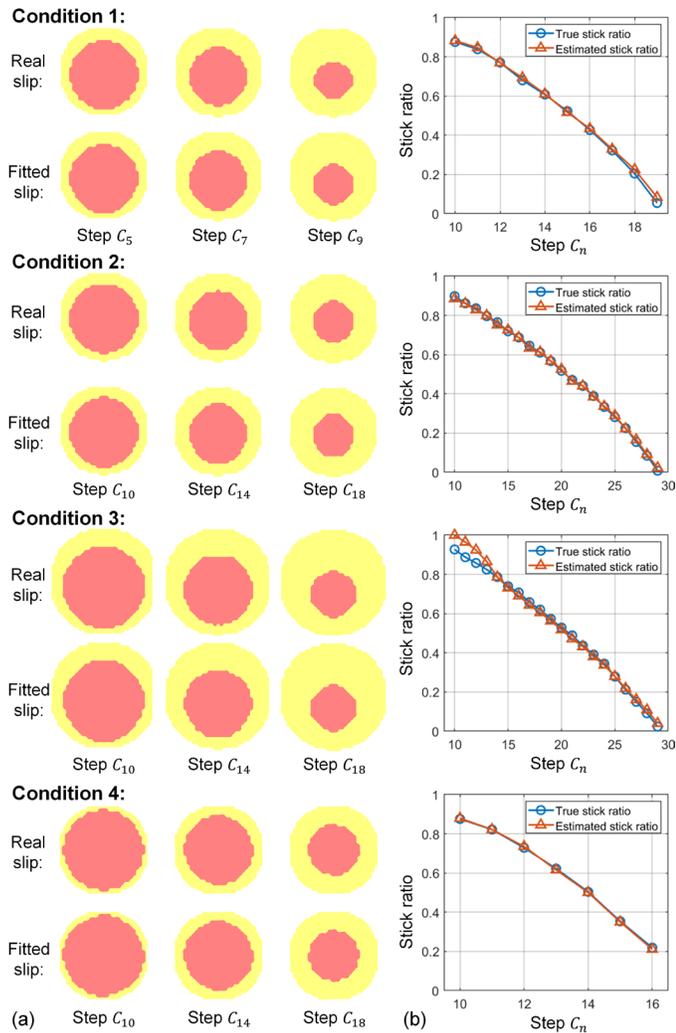

**Fig. 8.** Evaluation of incipient slip detection in the spatial dimension. (a) The distribution of real stick-slip regions under four different conditions, compared with the segmentation results obtained using the proposed method. Three analysis steps were selected for each condition. (b) The real stick-slip ratio under the four conditions compared with the estimated stick-slip ratio obtained from incipient slip detection.

being consistently zero.

Furthermore, the stick-slip ratio was used to quantitatively evaluate the degree of incipient slip, as shown in Fig. 8(b). This metric, which has been repeatedly used in slip detection research, was defined as the area of the stick region $S_c$ divided by the area of the entire contact region $S_a$:

$$cf = f_t/\mu f_n, \tag{37}$$

The range of $sr$ is $[0,1]$. The smaller the value of $sr$, the smaller the area of the stick region, indicating that the contact state is closer to the macro slip. In this way, the degree of incipient slip can be quantitatively described, and the accuracy of different methods can be evaluated. The results in Fig. 8(b) show that, for Conditions 1, 2, and 4, the proposed method still estimates the spatial distribution of incipient slip well. For Condition 3, at smaller analysis steps, there is a more significant deviation between the estimated stick-slip ratio and the actual one. The reason is that, at this stage, the response of the C-Strain rate is smaller [see Fig. 6(f)], and the area of the slip region is smaller. Thus, the measurement noise is closer to the measured values (i.e., the relative error is larger). Overall, the effectiveness of the proposed method was validated.

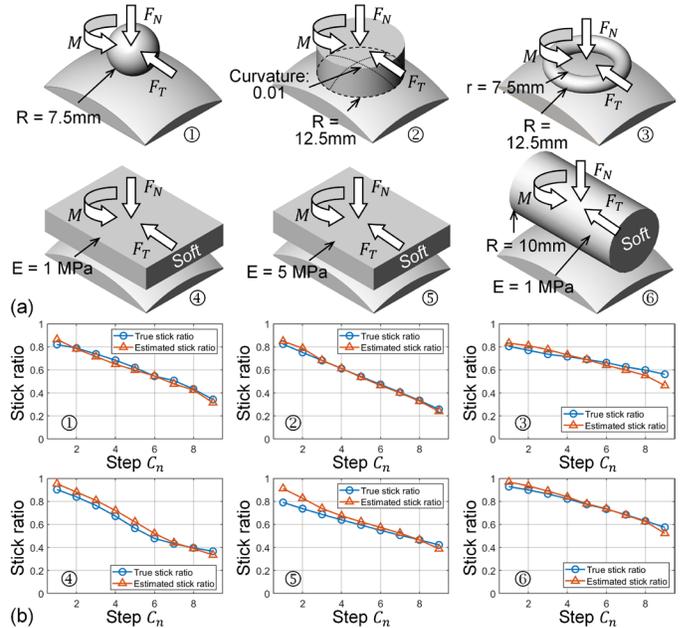

**Fig. 9.** Validation of adaptability to shape, stiffness, and combined loading. (a) Six objects with different geometries and elastic moduli, subjected to both tangential force and torque during contact. (b) Real stick-slip ratio and estimated stick-slip ratio under these conditions.

### D. Applicability Validation

1) *Object adaptability*: The applicability of the proposed slip detection method was validated for contact objects with different shapes and stiffness, as well as for combined loading conditions involving both tangential force and torque. The simulation models used are shown in Fig. 9(a). Test objects 1 to 3 (rigid objects) are a spherical object, a spherical object with local curvature smaller than the elastomer's surface, and a ring-shaped object, respectively. Test objects 4 to 5 (flexible objects) are square objects with elastic moduli of 1 MPa and 5 MPa, respectively, while test object 6 simultaneously changes both the shape and stiffness of the object. A normal force of $F_N = 50N$ was first applied, followed by a gradual application of a tangential force $F_T = 12N$ along the x-axis and a torque $M = 60N \cdot mm$ along the -z axis. Both the tangential force and torque increased uniformly over ten analysis steps, from $C_0$ to $C_9$.

Starting from the first frame when the tangential load begins to be applied, Fig. 9(b) shows the stick-slip ratio from step $C_1$ to $C_9$ for the six cases mentioned above. As the tangential force and torque increase, the stick-slip ratio monotonically decreases. For objects with different shapes and curvatures, the estimated stick-slip ratio aligns with the ground truth. Since it relies solely on local information, the proposed method does not require the assumption of a stick center region and can handle cases where the contact area is hollow. Additionally, despite being based on simplified models including Hertzian contact and virtual elastic beams, the method is also applicable to soft contact objects without requiring threshold segmentation of the deformation field [27]. In summary, the proposed incipient slip detection method relies less on measurement techniques and prior information compared to existing methods.

*2) Working condition adaptability*: The proposed method was then compared with three high-resolution incipient slip detection methods [23], [26], and [27]. Since spatial characteristics of incipient slip were considered, some of them could be applicable to objects of different shapes, while others addressed the problem of detecting soft objects. However, due to the lack of perception of temporal characteristics, these methods required constraints on loading and motion (e.g., constant grasp force or 2-D rigid body motion). Thus, such approaches could not handle varying normal forces or reversing tangential forces [16]. These limitations make it challenging to apply such methods in grasping and manipulation. Thus, we primarily compared the applicability of different methods to varying working conditions.

The scenario shown in Fig. 10(a) was analyzed using the simulation model in Fig. 5(a): Based on Condition 1, after the tangential force is fully applied, the normal force is further increased to $F_N = 20N$. After the normal force is fully loaded, the tangential force is increased to $2F_T = 15.9N$. Fig. 10(b) shows the measurement results at the reference point $p_c$ in Fig. 7(a) during this process, including the C-Strain rate and the estimated local slip state based on the strain-rate extreme events. The strain-rate peak caused by normal contact is excluded according to the description in Section II-D. During this process, when the normal force is increased for the second time. Since the contact surface returns to a stable state, $p_c$ returns to a non-slip state. When the tangential force is increased for the second time, $p_c$ undergoes local slip again. Due to the residual stress and strain from the previous slip, the corresponding C-Strain rate is more pronounced during the second slip than the first time.

Fig. 10(c) and (d) compare the stick-slip region segmentation and stick-slip ratio results obtained using different methods, respectively. For the method in [23], since the details of the thresholds for the region filtering were not provided, a similar proportion was selected according to [26] for reproduction. This method estimates the slip distribution by fitting rigid body motion to points assumed to be in the stick region. At the initial stage of loading, the stick region is sufficiently large, and all selected reference points are located within the stick region. However, as the stick region decreases, the fitting error gradually increases. This is reflected in the stick-slip ratio measurement results. From the second increase in normal and tangential forces onward, the residual strain affects the displacement field's non-uniformity within the assumed stick region, leading to a failure in fitting the slip field. For example, at Step $C_{24}$ in Fig. 10(c), some of the selected reference points have already slipped, resulting in significant errors in region segmentation.

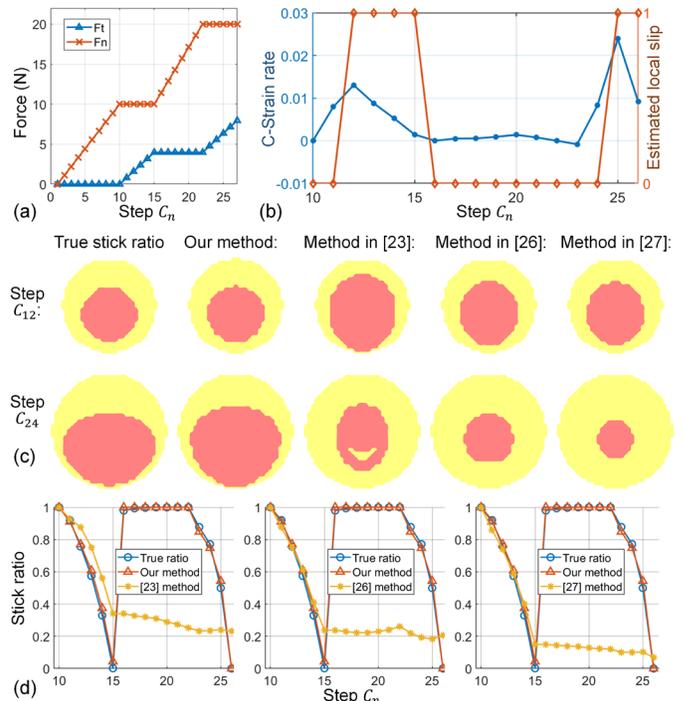

**Fig. 10.** Comparison of the applicability of different methods for the case of change in normal force. (a) Variation of loading forces with analysis steps. (b) Variation of the C-Strain rate and the estimated local slip state of $p_c$ in Fig. 7(a). (c) Distribution of the real stick-slip region and the segmented result obtained by the proposed method. (d) Comparison of stick-slip ratio between the proposed method and three existing methods. Benchmark: real stick-slip ratio. Baseline: methods in [23], [26], and [27].

The methods in [26] and [27] consider local deformation degree and deformation field gradient, respectively. Their core idea is to determine the stick region by thresholding the deformation field or comparing local differences. These methods effectively fit the slip field during the first tangential loading process, as shown at Step $C_{12}$ in Fig. 10(c). However, when the normal force increases for the second time, the deformation caused by the first tangential loading remains, and the overall change is insignificant. As a result, both methods incorrectly assess the current contact stability. They fail for detection thereafter due to the inability to distinguish the newly added slip region from the existing distribution of stick-slip regions.

In summary, the methods mentioned above are not suitable for contact processes with repeated loading. When a point on the contact surface undergoes multiple state transitions from slip to stability, existing approaches cannot effectively distinguish the influence of residual stress on the deformation field, leading to failure. However, in practical grasping processes, the manipulator always adjusts the grasping force and posture continuously. Therefore, the contact surface will inevitably undergo repeated changes in the incipient slip forward and reverse direction, making it difficult for existing methods to be practical. In contrast, the proposed method aligns well with the actual contact state even during the second variations of normal and tangential forces. Section IV-B will also show its adaptation to reverse tangential loading situations. By modeling the temporal characteristics of the C-Strain rate, the defined slip detection strategy does not rely on the overall distribution of deformation or historical information. Even in the presence of residual strain, the extreme events of the C-Strain rate can accurately reflect the variation of newly added effects, thereby capturing the spatial characteristics of stick-slip state changes.

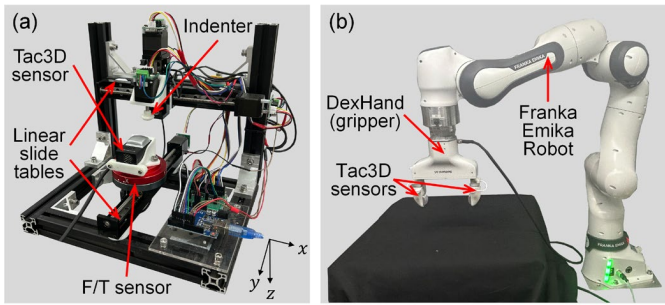

**Fig. 11.** Experimental setup. (a) Static measurement evaluation. (b) Online measurement evaluation.

## IV. EXPERIMENT

### A. Experimental Setup

In this section, two sets of experiments were performed: static experiments on a calibration platform and online experiments on a robotic arm system. Fig. 11(a) shows the experimental setup for static evaluation, configured according to the design in [43]. The linear slide tables (with a resolution of 0.1 mm) control the tangential and normal contact between an indenter and the Tac3D sensor. In contrast, the equipped F/T sensor provides 3-D force information on the contact surface (with a resolution of 0.01 N). The online experimental platform is shown in Fig. 11(b). We mounted a custom-built DexHand robotic hand (parallel gripper) on a Franka Emika robot and installed two Tac3D tactile sensors on the gripper. The Tac3D sensor records tactile data at 30 frames per second, with an effective measurement grid of 20×20 points. These two metrics represent the temporal and spatial resolution of the sensor in detecting incipient slip fields.

### B. Static Measurement Evaluation

1) *Spatial characteristics*: On the platform shown in Fig. 11(a), tangential contact conditions were constructed to analyze the strain process before macro slip. Fig. 13(a) illustrates the variation of concentrated contact forces in one set of experiments (with a tangential movement speed of 0.2 mm/s). First, normal contact was applied along the z-axis. Then, a constant-speed tangential movement was applied in the x- axis. In the third stage, reverse movement at a constant speed was applied in the negative x- axis. Finally, the indenter was released in the negative z-axis, completing one cycle of contact. By varying different tangential movement speeds (0.2, 0.4, 0.6, 0.8, and 1.0 mm/s), the factors influencing the C-Strain rate were investigated. The above process was repeated 10 times for each condition.

Fig. 12 shows the evolution of distribution forces, C-Strain rate fields, and estimated stick-slip region segmentation during the tangential loading process. The distribution forces were calculated using the inverse finite-element method [44], [45], [46]. During shearing along the x-direction, the stick region gradually shrinks from the outside inward (see the first three columns), which is consistent with the simulation results in Section III. When shear is reversed, the tangential load first decreases and then increases. As a result, the contact surface momentarily returns to a full stick state (see the fourth column), and the evolution process of the slip region repeats again. The difference lies in the asymmetric deformation of the sensor's elastomer due to contact (see Fig. 12(a)). This issue causes the stick region to shift in the opposite direction relative to the contact area's center, corresponding to the current tangential force direction. Fig. 12(c) results show that the stick-slip region distribution estimated using the C-Strain rate is consistent with the above observations. Therefore, the proposed detection method can effectively capture the spatial characteristics of incipient slip.

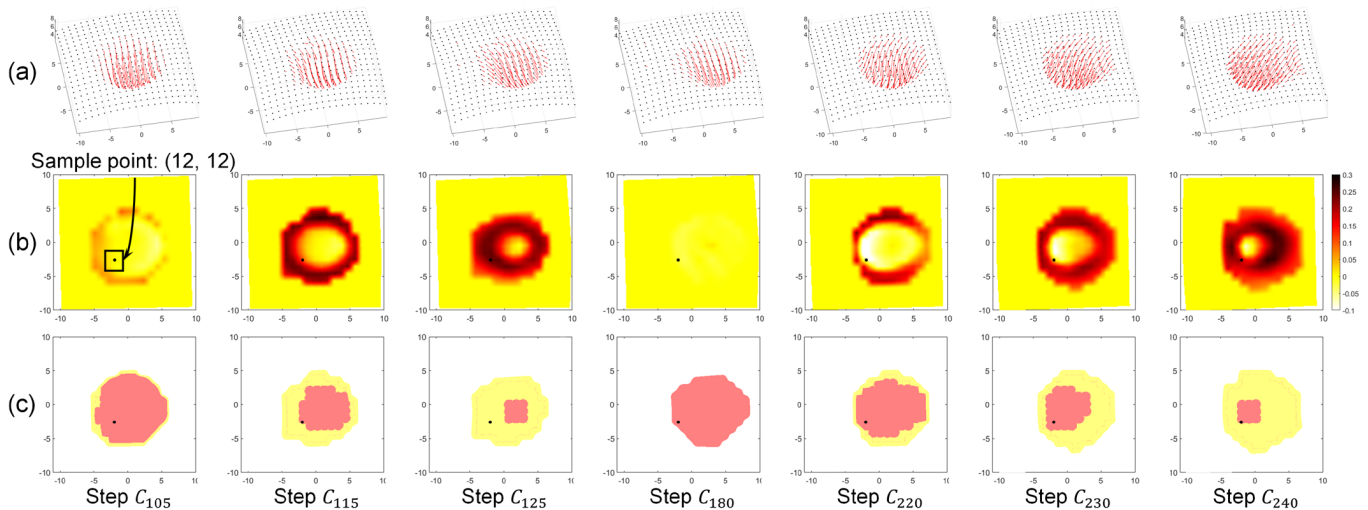

**Fig. 12.** Evolution of distribution fields during tangential loading and reverse loading. (a) 3-D contact deformation and distributed forces. (b) C-Strain rate field. (c) Estimated stick-slip region segmentation.

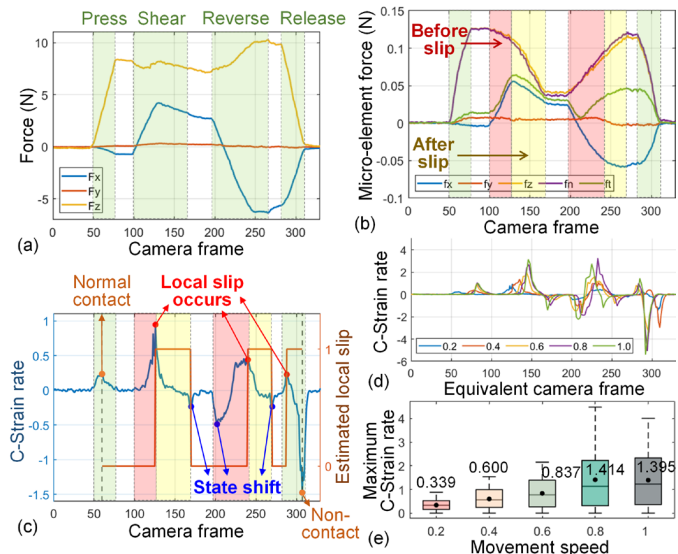

**Fig. 13.** Static measurement evaluation. (a) Temporal evolution of contact forces, including four stages: press, shear, reverse, and release. (b) Variation of micro-element forces at the reference point in Fig. 12(b). The slope change of the micro-element tangential force distinguishes the boundary before and after the slip occurs. (c) Temporal evolution of the C-Strain rate and the estimated local slip at the reference point. The extreme events of the C-Strain rate correspond to four states: normal contact, onset of local slip, state transition, and non-contact. (d) Comparison of the C-Strain rate at the reference point under five different speeds. For consistency, the horizontal axis is scaled according to the tangential speed. (e) Comparison of the maximum C-Strain rate under five different speeds. The box plots represent the discrete results obtained at different reference points, while the black dots and numbers indicate the statistical mean values.

*2) Temporal characteristics:* In Fig. 12, reference point (12, 12) was selected to analyze the evolution of local contact state. Fig. 13(b) and 13(c) show the evolution of the micro-element force and the C-Strain rate at the reference point, corresponding to the same loading process as shown in Fig. 13(a). Specifically, the micro-element normal force and tangential force were calculated using the local normal vector $n$ of the contact surface according to Eq. (32):

$$f_n = -(f \cdot n) \cdot n, \tag{38}$$

$$f_t = -f + (f \cdot n) \cdot n, \tag{39}$$

where $f$ represents the 3-D elemental force.

Fig. 13(b) shows that the evolution of the micro-element force also corresponds to the four stages of the concentrated force variation. Meanwhile, it reflects the influence of local slip during the two tangential contact processes: when local slip occurs, the slope of the micro-element normal force does not change significantly, but the slope of the micro-element tangential force decreases markedly. Therefore, the occurrence of local slip can be determined based on the variation in the micro-element force. Note that the ratio of the micro-element tangential force to the normal force differs significantly between the two instances of local slip. According to Coulomb's law of friction, the force ratio at the onset of slip should only depend on the friction coefficient and the shear failure limit. On one hand, the friction coefficient may vary during the contact process. On the other hand, the simplified friction model cannot accurately explain the actual force distribution. These results suggest that the reliability of incipient slip detection methods relying on a priori information (like the Coulomb's friction coefficient) is questionable under complex contact conditions.

The four loading stages and two slip events identified in Fig. 13(b) (highlighted by different colored regions) are plotted in Fig. 13(c). It can be observed that the strain-rate extreme events occur precisely at the boundaries between different stages:

1) When the object contacts the sensor, the first peak in the C-Strain rate occurs, indicating the initiation of a local non-slip contact state and the beginning of slip detection.

2) During the first tangential loading process, the second peak signifies the occurrence of local slip. As the tangential force ceases to increase and starts to decrease, internal stress redistributes until the micro-element reaches a stable state again [27]. At this moment, the C-Strain rate reaches the first negative peak.

3) The reference point then experiences the reverse strain release process explained in Fig. 3(d), while the contact surface remains in a stable state. Similarly, at the end of the second tangential loading, the C-Strain rate reaches the second negative peak.

4) As the normal contact is released, residual stress drives the reference point to move instantaneously along the tangential direction relative to the contact surface. It gradually separates from the contact surface, resulting in the third instance of local slip.

5) Finally, the C-Strain rate reaches its minimum value as the contact strain is fully released.

Therefore, in the above process, the local slip estimated based on C-Strain rate extreme events coincides with the segmentation of different stages identified in Fig. 13(b). This indicates that the proposed detection method effectively characterizes the temporal dimension of incipient slip. In addition, for the C-Strain rate extreme events corresponding to the initiation and termination of contact, it is possible to distinguish them from the extreme events due to slip and state switching by the change in the normal force or displacement of the microelement. In summary, the proposed method is minimally affected by changes in friction conditions, and can still respond to changes in local slip states under complex working conditions.

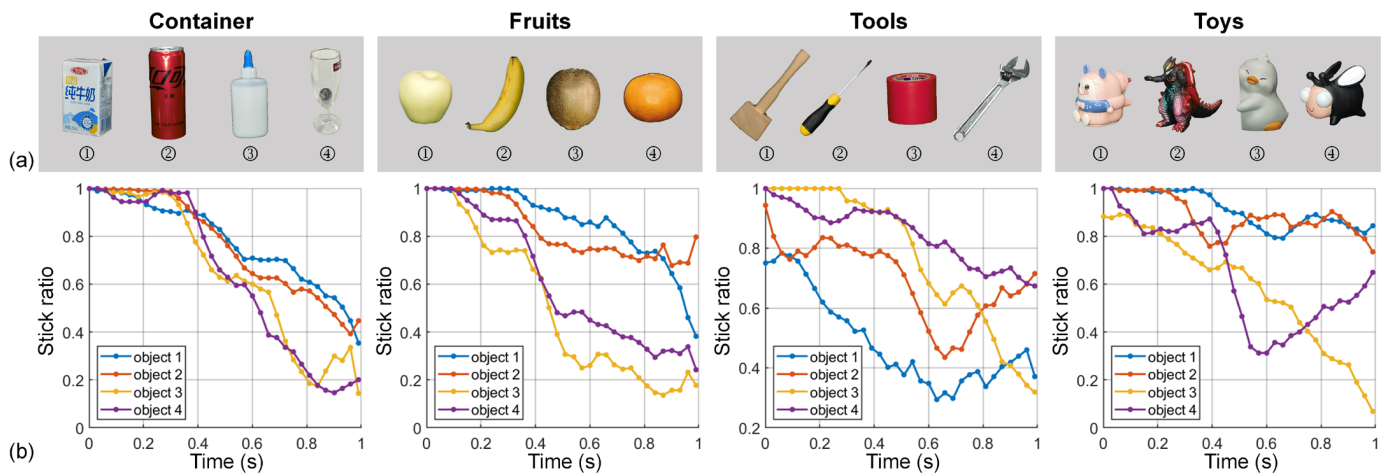

**Fig. 14.** Evaluation of different contact objects. (a) Test objects of four groups. (b) Variation of the estimated stick-slip ratio during the lifting process for different objects.

Further analysis was conducted on the impact of tangential motion speed on incipient slip detection. Fig. 13(d) shows the evolution of the C-Strain rate at the reference point under five different speeds. The x-axis represents the equivalent sensing frame scaled according to tangential velocity for aligning the time-series signals for a more intuitive comparison across different speeds. The results indicate that the evolution of the C-Strain rate is similar across different tangential speeds, with differences mainly observed in the size and phase of the extrema. The statistical analysis of the maximum C-Strain rate at each reference point across all test cases is shown in Fig. 13(e). Within the range of 0.2 to 0.8, as the tangential speed increases, the response of the C-Strain rate becomes more pronounced. But when the speed increases further, there is no significant change in the response. In other words, there should be an upper limit to the magnitude of the C-Strain rate extrema under the same normal force. This characteristic is likely related to the friction coefficient and the softness or stiffness of the contact surface. Overall, beyond normal force and friction coefficient, the C-Strain rate also implicitly includes characteristics of the shear speed. A deeper understanding of the feature strain rate will facilitate the development of online detection methods for tactile features such as normal force, friction coefficient, and shear speed.

### C. On-Line Measurement Evaluation

Based on the system shown in Fig. 11(b), the performance of the proposed method in actual robot grasping tasks under unstructured scenarios was studied. In this section, in addition to adapting to different contact properties and changes in load, another main challenge of this task lies in requiring the proposed method to respond within a very short time (about 1 second), whereas previous works [16],[26]-[27] required over 30 seconds of measurement time to obtain effective results.

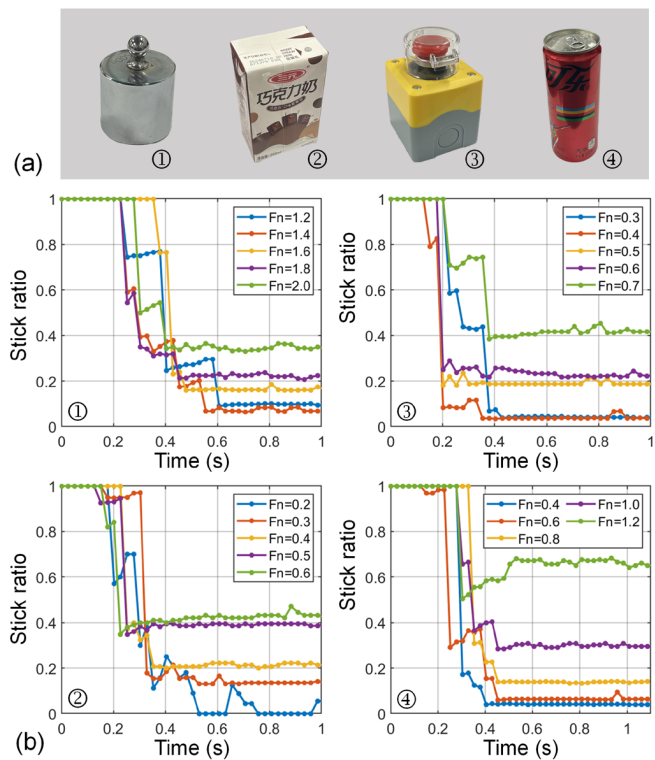

**Fig. 15.** Evaluation of different load forces. (a) Test objects. (b) Variation of the estimated stick-slip ratio during the lifting process under different grasp forces. During the process, the robot starts lifting the object at 0.2s with a speed of 3 mm/s, lifts for 0.6s, and then remains stationary for 0.2s.

*1) Evaluation of different contact objects:* As shown in Fig.

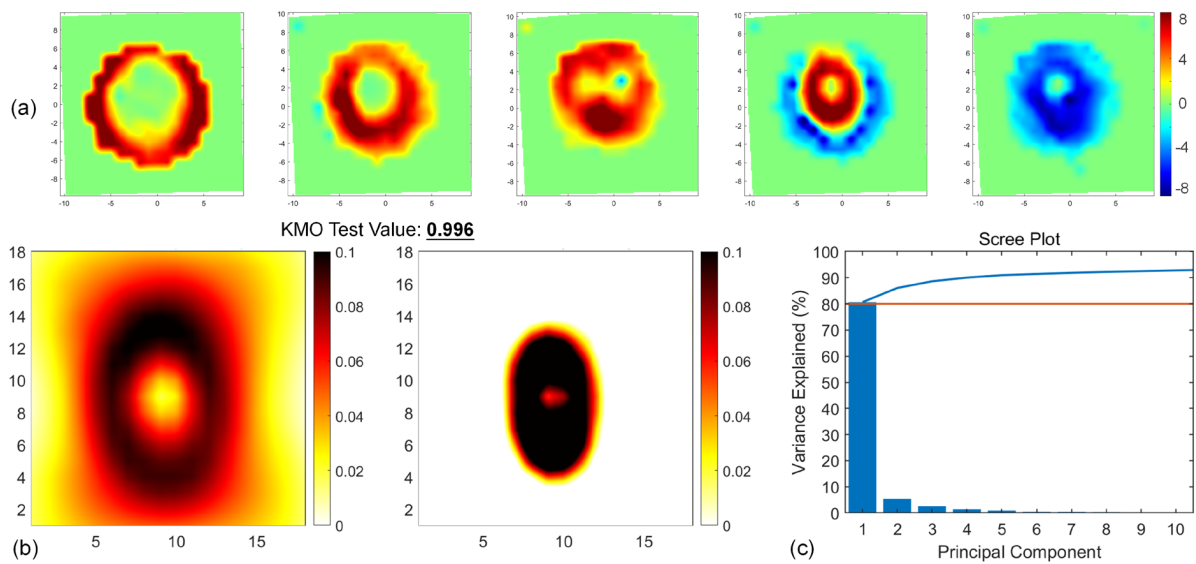

**Fig. 16.** Principal component analysis of C-Strain rate distribution. (a) Distribution of C-Strain fields. (b) The first two principal components of the C-Strain field field. (c) Percentage of explained variance scores for different principal components. Histogram: Variance interpretation rate. Blue line: cumulative explainable variance contribution curve. Orange line: cumulative contribution ratio of 80%.

14(a), 12 household objects from four categories were selected. For the Containers category, the objects contained internal fillings, causing the center of gravity to shift during the lifting process. Test objects in the Fruits category were relatively soft with smooth surfaces. The Tools category involved both translational and rotational slip, with significant differences in the properties of the contact surfaces. Objects in the Toys category had irregular geometries and considerable differences in weight (for example, object 3 was five times heavier than object 4). The experimental procedure was as follows: starting from the moment the gripper made contact with the object, tactile information was recorded by the Tac3D sensor. For each object, pre-experiments were conducted to determine a suitable target grasp force—strong enough to lift the object but not excessive (about 1.5 to 2 times the minimum required grasp force). Once the grasp force reached the target value, the object was lifted at a speed of 1 mm/s for 2 seconds.

The stick-slip ratios were calculated during the loading process of each experiment, as shown in Fig. 14(b). Regardless of contact conditions such as surface material, curvature, or distribution continuity, the proposed method enables online estimation of the degree of incipient slip, consistent with the progression of the loading process. Under different conditions, the variations in the stick-slip ratio reflect the distinct properties of the objects: the smoother the object surface, the faster the decline in the stick-slip ratio. In addition, since the method relies on the C-Strain rate, its response must be sufficiently significant to be detected. Therefore, for lighter objects (e.g., object 2 in the Tools category) and complex objects with complex geometry (e.g., objects 2 and 4 in the Toys category), the measurement results showed relatively lower confidence levels. Overall, the proposed method proves applicable to objects with different shapes, materials, and weights.

*2) Evaluation of different load forces*: Grasping and lifting operations were performed on the four objects shown in Fig. 15(a). For each object, five different grasp forces were selected, and 10 tests were conducted for each condition. Each test contained approximately 200 frames of 20×20 C-Strain rate fields. Fig. 15(b) shows the variation of the stick-slip ratio calculated using the proposed method during the lifting process (one example was selected for each condition). The variation of the stick-slip ratio is consistent with the characteristics of the loading process: the smaller the grasp force, the faster the stick-slip ratio decreases, and the lower the value at the end of the task. When the grasp force is below a certain level, the stick-slip ratio rapidly approaches zero, indicating that macro slip has occurred. Except for object 2, this value is not exactly zero, mainly due to the thresholding technique used for noise reduction (see Section II-D).

Additionally, for all four objects, the stick-slip ratio shows a distinct staircase pattern. For the same drop, multiple stages of decrease are observed under certain conditions, which can be explained by the two-stage drop behavior of stick-slip stress drops [35]. These results indicate that the proposed method is suitable for online and rapid incipient slip detection in grasping applications, showing significant improvements in real-time performance compared to existing stick-slip ratio measurement methods.

*3) Evaluation of repeatability*: Fig. 16(a) presents the distribution of the C-Strain field during an extreme event response, selected from one grasp-and-lift operation for object 3 in Fig 15(a). As the tangential force increases, the degree of incipient slip gradually increases, and the C-Strain rate wave propagates from the outer side to the inner side. Whenever the normal force is adjusted to prevent macro slip, the stability of the contact surface is improved. At this moment, stress near the edge of the contact region is released and returns to a local stick state, showing a negative C-Strain rate. This observation aligns with the theoretical explanation in Section II and the simulation analysis in Section III. During the coordinated adjustment of tangential and normal forces, this process repeatedly occurs. Therefore, under complex loading conditions, the evolution of incipient slip is no longer the idealized situation described in Fig. 1, as residual stress and strain affect the distribution of the slip region and stick region. The proposed detection method is less influenced by the deformation history, making it more applicable than existing methods relying on deformation and tactile images.

After excluding the non-contact phase and instances with no significant response in the C-Strain rate field, a total of 1635 sets of tactile data corresponding to strain extreme events were obtained. Principal component analysis (PCA) was conducted to investigate the typical patterns of the C-Strain rate's spatial distribution. A sampling adequacy test is performed to evaluate the correlation among variables in the multivariate statistics. The KMO (Kaiser-Meyer-Olkin) test statistic is 0.996, indicating good suitability of the characteristic strain field data for PCA.

Fig. 16(b) presents the first two principal components calculated using the PCA function in MATLAB. The results demonstrate consistency between theory and practice: the first principal component resembles the C-Strain rate shown in Fig. 2(c) and Fig. 6(d)-(g), representing the influence of tangential force variations. The second principal component reflects a nearly uniform C-Strain rate distribution throughout the contact region except for the central part, which can be interpreted as the contribution of normal load variations. When approaching the center, the tangential strain caused by normal compression is relatively small.

Additionally, the explained variance percentage of different principal components is compared, as shown in Fig. 16(c). A higher variance percentage of a principal component indicates more information contained in that feature. Statistically, the first principal component accounts for the largest proportion, reaching 80%. If only the first two principal components are used for dimensionality reduction, approximately 85% of the effective information can still be retained. These results indicate that the evolution pattern of the described C-Strain rate, influenced by tangential and normal forces, is universal and can explain the incipient slip phenomenon observed in actual grasping tasks.

## V. Application

In this section, we preliminarily explore the adaptability of the proposed method in different tasks of friction recognition and grasping control.

### A. Friction Recognition

Reference [47] emphasized the importance of accurate friction models for robotic applications. The authors proposed a method that modeled the friction property as random variables influenced by the normal force $F_n$ and contact velocity $v$. A simple approach involved fitting the friction coefficient using a linear regression model:

$$\mu = \beta_0 + \beta_1 v + \beta_2 f_n + \varepsilon, \qquad (40)$$

where $v$ denotes the tangential velocity, $f_n$ denotes the normal force, and $\varepsilon$ represents the stochastic variation in $\mu$.

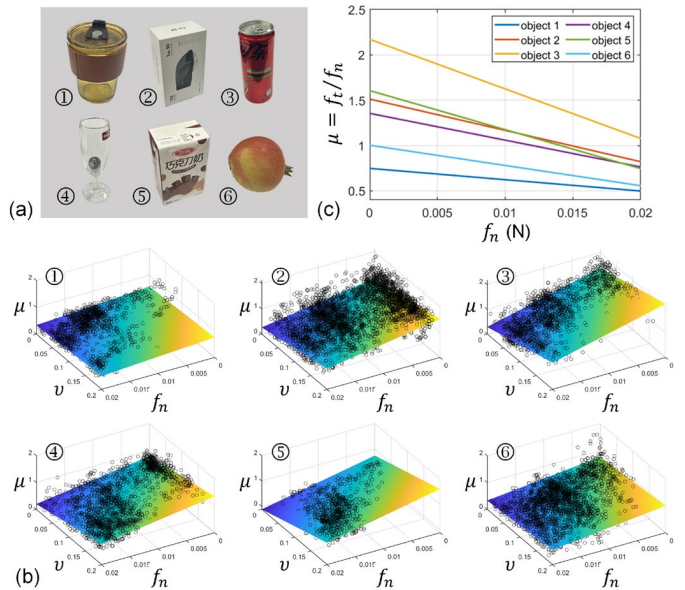

**Fig. 17.** Friction recognition based on incipient slip detection. (a) Test objects. (b) Distribution of random friction measurement results for six objects. The plane represents the fitted linear friction model. (c) Projection of the friction models on the $v = 0$ plane.

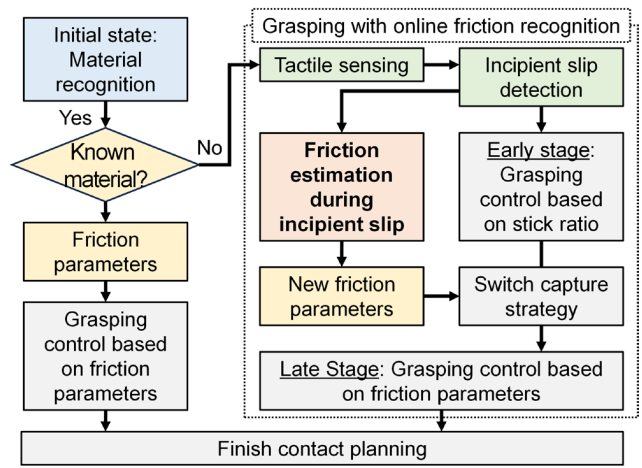

**Fig. 18.** Grasping control based on online friction recognition, inspired by [47].

Our work provides a local slip detection approach independent of friction and distributed force measurements. Its significance lies in enabling the online estimation of friction parameters for practical applications. This can be achieved by statistically analyzing the micro-element velocity and force at sampling points where local slip is detected, and fitting the coefficients according to the selected friction model. Without loss of generality, we used the linear model in Eq. (40) to describe the friction property. Six objects made of different materials were selected, as shown in Fig. 17(a), and tactile data were collected online during the object lifting process using the control method from [46]. Fig. 17(b) shows the random friction distributions and the fitted linear friction models for each material, and the projections of the friction models on the $v = 0$ plane are shown in Fig. 17(c).

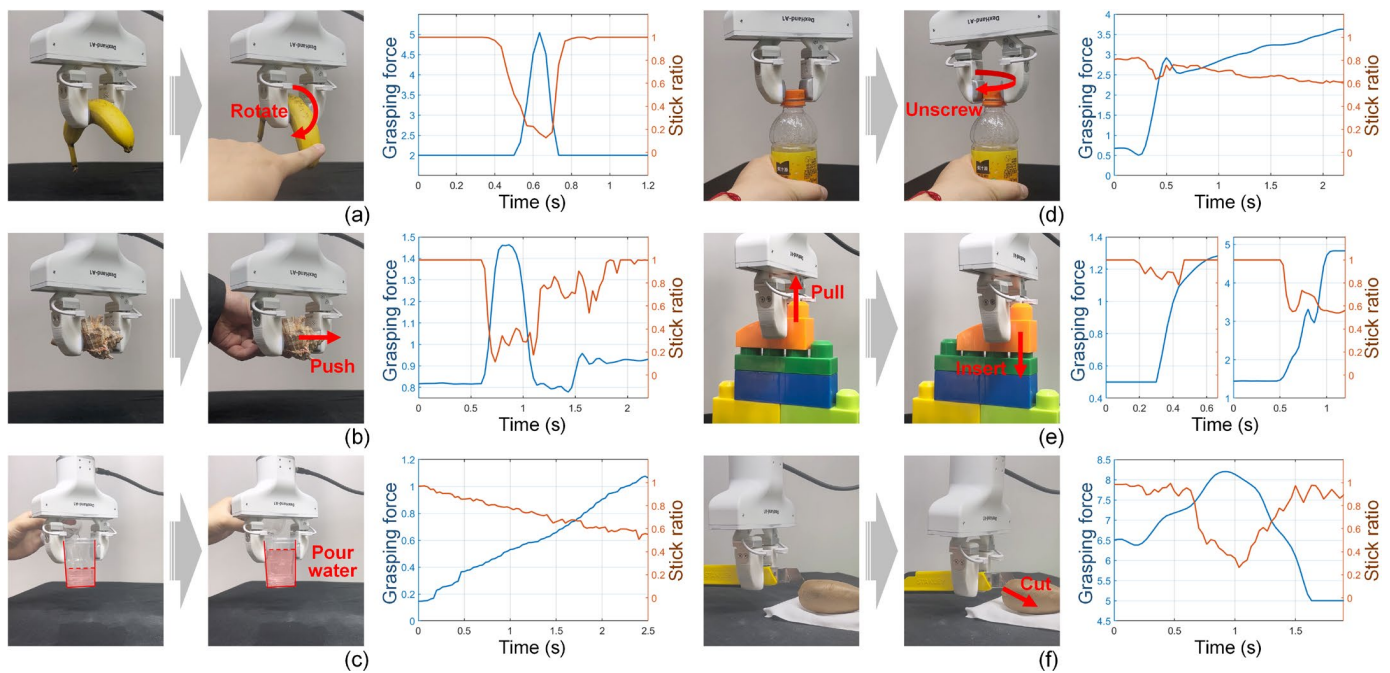

**Fig. 19.** Applications in grasping and manipulation. For each set of subfigures, the left side shows the task scenario, while the right side illustrates the variation of stick-slip ratio and grasp force over time during task execution. (a)-(c) Grasping tasks. (a) Banana; (b) Sea shell; (c) Plastic cup being filled with water. (d)-(f) Manipulation tasks. (d) Twisting a bottle cap; (e) Building blocks; (f) Cutting a kiwifruit.

The results indicate that the constructed friction parameters can describe the friction characteristics of different materials. For objects 2 and 5, whose contact surfaces were similar in shape and were both made of cardboard, the obtained friction characteristics were also similar. There is significant stochastic variation among different objects, and objects with larger $\varepsilon$ values (e.g., objects 2 and 6) require a wider range of samples to estimate the friction model parameters accurately. It is noteworthy that, since no fixed calibration procedure is required, the measurement of friction parameters does not interfere with the execution of the controlling but can be performed simultaneously. Therefore, the online estimation of friction parameters based on incipient slip detection can distinguish the friction characteristics of objects even during the grasping and manipulation tasks.

Based on the above analysis, we recommend a robot control framework based on friction parameter estimation, as shown in Fig. 18. First, a material recognition mechanism is used to identify the type of object; then, friction parameters of the object are estimated using a regression model, either by querying a material database or through on-site measurements.

Finally, the estimated parameters are applied to the actual control of grasping and manipulation. This framework is similar to the approach proposed in [47]. The difference lies in the fact that incipient slip detection based on strain-rate extreme events enables real-time friction parameter estimation during operation. In other words, the proposed method does not require inducing macro slip between the object and the robot's fingertip to obtain random friction parameters during movement. Instead, it allows measurement through local slip while ensuring that the object remains securely grasped. To ensure stability throughout the process, grasp force can be controlled using characteristics such as stick-slip ratio in the early stages when friction parameters are not yet determined. Once the friction parameters are successfully estimated, the control strategy can be updated accordingly. Overall, local slip sensing minimizes the need for a specific loading process during the detection and enhances the practicality of the described framework.

### B. Grasping Control

Perception of slip and friction is crucial for accurately identifying object properties and performing reliable grasps, especially in unstructured environments. We demonstrate three sets of grasping tasks and three sets of manipulation tasks, as shown in Fig. 19, to illustrate the role of the proposed incipient slip detection method in enhancing task adaptability.

*1) Grasping tasks:* Three sets of grasping control experiments were conducted: banana, sea shell, and a gradually filling cup of water. Our objective was not merely to successfully grasp the objects but to investigate whether the grasp force can be effectively adjusted to prevent slippage when external loads change rapidly. In the banana grasping task [see Fig. 19(a)], a rapid twist was manually applied to provide perturbation. As the degree of rotational slip increased (indicated by a decrease in stick-slip ratio), the grasp force was quickly adjusted to ensure stability.

In Fig. 19(b), the object of interest was a sea shell with an irregular geometric shape. Despite the complex contact conditions, effective estimation of the temporal characteristics of incipient slip enabled the gripper to respond promptly to the applied tangential load. In addition to responding to rapid slip trends, the proposed method was also effective for gradually changing tangential loads.

As shown in Fig. 19(c), when water was gradually poured into the plastic cup, the gravitational force increased continuously. During this process, effective measurements of incipient slip allowed the gripper to simultaneously increase the grasp force, slowing down the decline in the stick-slip ratio. The process of increasing gravity was coordinated with the increase in grasp force.

2) *Manipulation tasks*: Based on the perception of incipient slip, we were able to complete three different manipulation tasks. When tightening a bottle cap using the robotic arm, the required torque increased as the cap was tightened [see Fig. 19(d)]. So, the gripper needed to increase the grasp force to prevent slippage. Initially, a small force was preset for the gripper. As the cap tightens, the area of the slip region increased. The detector perceived the spatial characteristics of incipient slip to prevent the object from slipping off. When a rapid decreased in the stick-slip ratio was detected, the gripper quickly increased the grasp force, and continued to increase it gradually to control the reduction of the stick-slip ratio.

Fig. 19(e) shows the manipulation task of building blocks. During the process of pulling out and inserting the yellow block, the friction between the green block below and the yellow block suddenly increased. This process is similar to peg assembly with added resistance. Perception of the contact state ensured that the blocks maintain stable contact with the sensor at all times. The difference in the change in stability between the front and back contact is also reflected in the end-state grasp force.

Finally, Fig. 19(f) shows an everyday task of cutting fruit. In this experiment, the toughness of the kiwifruit itself and the stickiness of its flesh pose resistance to the knife edge, especially when cutting through the skin, requiring a significant force to break through. Without proper grasp force control during this process, the knife would be unable to penetrate the flesh, resulting in unwanted in-hand pivoting. Tactile sensing was used to detect both translational and rotational slips in order to determine the necessary grasp force. The robotic arm controlled the knife to penetrate the kiwi until the sensor detected the localized slip caused by the contact. Based on this information, the grasp force was adjusted. Eventually, once the knife was fully inserted, the kiwi provided normal force support, which reduced the translational slip due to gravity. As a result, the final grasp force is smaller than the force before cutting, and the grasp control process aligns with the expectations for this task.

## VI. Conclusion

This article presents a modeling and detection method for the spatial-temporal characteristics of incipient slip detection, and also discusses its application in robot recognition and control. Based on the constructed contact model, the extreme events of the characteristic strain rate are identified as effective features for characterizing local slip, containing both spatial and temporal dimensions of complete information. According to the experimental results, the defined features numerically characterize the friction strength and contact velocity, while the extreme events describe the localized slip state. Compared to existing methods, this approach has the advantage of adaptability in object types and working conditions. Requiring no prior knowledge or mechanical measurements, it is particularly applicable to complex contact conditions with residual stresses, such as varying normal forces and the multi-directional slip. Moreover, the method can be applied under rapid changes in contact force and has the advantage of fast response. The effective application in friction estimation and grasping tasks demonstrates the potential of this method to expand the capabilities of robotic contact perception and planning control.

The limitation of this method lies in its inability to apply to objects with particularly small friction coefficients or scenarios with very low normal force. In such cases, the absolute value of the characteristic strain rate is very small, making it difficult to distinguish it from purely normal contact situations. This phenomenon is also present in human tactile behavior: as the contact surface becomes smoother, it becomes more challenging to regulate the force [3]-[4]. Therefore, features beyond tactile information (such as visual images) and prior knowledge are necessary in the follow-up work.

Furthermore, this article presents a traditional detection method based on physical modeling. Given that experiments have already shown that the characteristic strain and strain rate distributions encompass multi-dimensional features, including normal force, friction characteristics, and tangential velocity, learning-based methods are expected to extract deeper tactile features, further enhancing contact perception capabilities. Future work will explore suitable network architectures for incipient slip detection and friction identification. In addition, grasp force regulation based on contact feature perception remains an open issue in robotics community. Since the grasping applications demonstrated still rely on control measurements designed for specific task characteristics, it is necessary to investigate how to better utilize slip and friction perception information to explore versatile control strategies.

## Appendix A

As shown in Fig. 2(a), the contact region is circular according to Hertzian contact theory and semi-elastic space approximation. Therefore, the normal force distribution and the normal force satisfy:

$$f_z = f_n \cdot \left[1 - \left(\frac{r}{r_a}\right)^2\right]^{1/2}, \quad (A1)$$

$$F_N = \frac{2}{3}\pi f_n r_a^2, \quad (A2)$$

where $f_n$ denotes the maximum distributed normal force and $r_a$ denotes the radius of the contact region. Let the radius of the stick region be $r_c$. According to Cattaneo's method [31], the analytical solution of the tangential force distribution of the localized slip problem can be constructed by combining the Hertzian stress distribution ($r = \sqrt{x^2 + y^2}$):

$$f_x = f_x^{(1)} - f_x^{(2)}, \quad (A3)$$

$$\text{where } f_x^{(1)} = f_{t1} \cdot \left[1 - \left(\frac{r}{r_a}\right)^2\right]^{1/2}, r \in [0, r_a], \quad (A4)$$

$$\text{and } f_x^{(2)} = f_{t2} \cdot \left[1 - \left(\frac{r}{r_c}\right)^2\right]^{1/2}, r \in [0, r_c]. \quad (A5)$$

Taking $f_x^{(2)}$ as an example, the tangential displacement field in the action region of $f_x^{(2)}$ can be calculated based on the potential theory according to Mindlin [31]:

$$u_x^{(2)}(r \le r_c) = \frac{\pi f_{t2}}{32 G r_c}\left[\begin{array}{c} 4(2-\nu)r_c^2 \\ -(4-3\nu)x^2 - (4-\nu)y^2 \end{array}\right], \quad (A6)$$

$$u_y^{(2)}(r \le r_c) = \frac{\pi f_{t2}}{32 G r_c} \cdot 2\nu xy, \quad (A7)$$

and the tangential displacement field outside the action region is derived from Illingworth's calculations [31]:

$$u_x^{(2)}(r > r_c) = \frac{f_{t2}}{8 G r_c} \cdot$$

$$\left\{\begin{array}{c} (2-\nu)\cdot\left[\begin{array}{c}\left(2r_c^2 - r^2\right)sin^{-1}\left(\frac{r_c}{r}\right)\\ + r_c r\left(1 - \frac{r_c^2}{r^2}\right)^{1/2}\end{array}\right]\\ +\frac{1}{2}\nu\cdot\left[\begin{array}{c}r^2 sin^{-1}\left(\frac{r_c}{r}\right)\\ +(2r_c^2 - r^2)\left(1 - \frac{r_c^2}{r^2}\right)\frac{r_c}{r}\end{array}\right]\cdot(x^2 - y^2)\end{array}\right\}, \quad (A8)$$

$$u_y^{(2)}(r > r_c) = \frac{f_{t2}\nu}{8 G r_c} \cdot \left[\begin{array}{c}r^2 sin^{-1}\left(\frac{r_c}{r}\right) + \\ (2r_c^2 - r^2)\left(1 - \frac{r_c^2}{r^2}\right)^{1/2}\frac{r_c}{r}\end{array}\right]xy. \quad (A9)$$

where $G$ denotes the shear modulus and $\nu$ represents the Poisson's ratio. Similarly, the tangential displacement field under the action of $f_x^{(1)}$ can be expressed as:

$$u_x^{(1)}(r \le r_a) = \frac{\pi f_{t1}}{32 G r_a}\cdot\left[\begin{array}{c}4(2-\nu)r_a^2\\ -(4-3\nu)x^2 - (4-\nu)y^2\end{array}\right], \quad (A10)$$

$$u_y^{(1)}(r \le r_a) = \frac{\pi f_{t1}}{32 G r_a}\cdot 2\nu xy. \quad (A11)$$

According to the superposition principle, the total contact deformation can be expressed as:

$$u_x = \begin{cases} u_x^{(1)}(r \le r_a) + u_x^{(2)}(r \le r_c), & r \in [0, r_c] \\ u_x^{(1)}(r \le r_a) + u_x^{(2)}(r > r_c), & r \in (r_c, r_a] \end{cases}, \quad (A12)$$

$$u_y = \begin{cases} u_y^{(1)}(r \le r_a) + u_y^{(2)}(r \le r_c), & r \in [0, r_c] \\ u_y^{(1)}(r \le r_a) + u_y^{(2)}(r > r_c), & r \in (r_c, r_a] \end{cases}. \quad (A13)$$

Besides, the stick region and slip region satisfy the geometric constraints and force constraints, respectively:

$$\begin{cases} u_x = \text{const and } u_y = \text{const}, r \in [0, r_c] \\ f_x = \mu f_z, r \in (r_c, r_a] \end{cases}, \quad (A14)$$

where $\mu$ is the coefficient of friction. Substituting Eqs. (A6)-(A11) into Eq. (A14) yields

$$u_x(r \le r_c) = \frac{(2-\nu)\pi\mu f_n}{8 G r_a}(r_a^2 - r_c^2), \quad (A15)$$

$$u_y(r \le r_c) = 0. \quad (A16)$$

The displacement of the stick region also represents the rigid body displacement. Thus, the slip at any position within the slip region can be determined by the difference between the displacement at that point and the rigid body displacement:

$$s_x(r > r_c) = u_x^{(1)}(r \le r_a)$$
$$+u_x^{(2)}(r > r_c) - u_x(r \le r_c), \quad (A17)$$

$$s_y(r > r_c) = u_y^{(1)}(r \le r_a)$$
$$+u_y^{(2)}(r > r_c) - u_y(r \le r_c), \quad (A18)$$

Substituting Eqs. (A6)-(A11), (A15)-(A16) into Eqs. (A17) and (A18). According to Johnson [31], $\nu/(4-2\nu)$ is of very small magnitude (about 0.09), so the small terms can be neglected. Therefore, the expression of the slip field is:

$$s_x(r > r_c) \approx$$

$$\frac{(2-\nu)\pi\mu f_n r_a}{8 G}\cdot\left\{\begin{array}{c}\left[1 - \frac{2}{\pi}sin^{-1}\left(\frac{r_c}{r}\right)\right]\left(1 - \frac{2r_c^2}{r^2}\right)\\ +\frac{2}{\pi}\frac{r_c}{r}\left(1 - \frac{r_c^2}{r^2}\right)^{1/2}\end{array}\right\}, \quad (A19)$$

$$s_y(r > r_c) \approx 0. \quad (A20)$$

## References

[1] R. S. Johansson, and J. R. Flanagan, "Coding and use of tactile signals from the fingertips in object manipulation tasks," *Nature Rev. Neurosci.*, vol. 10, no. 5, pp. 345–359, 2009.

[2] H. Khamis, S. J. Redmond, V. Macefield, and I. Birznieks, "Classification of texture and frictional condition at initial contact by tactile afferent responses," in *Proc. Int. Conf. Hum. Haptic Sens. Touch Enabled Comput. Appl.*, 2014, pp. 460–468.

[3] F. Schiltz, B. P. Delhaye, J.-L. Thonnard, and P. Lefèvre, "Grip force is adjusted at a level that maintains an upper bound on partial slip across friction conditions during object manipulation," *IEEE Trans. Haptics*, vol. 15, no. 1, pp. 2–7, 2021.

[4] B. P. Delhaye, F. Schiltz, F. Crevecoeur, J.-L. Thonnard and P. Lefèvre, "Fast grip force adaptation to friction relies on localized fingerpad strains," *Sci. Adv.*, vol. 10, no. 3, pp. 1-11, Jan. 2024.

[5] H. Khamis et al., "Friction sensing mechanisms for perception and motor control: Passive touch without sliding may not provide perceivable frictional information," *J. Neurophysiol.*, vol. 125, no. 3, pp. 809-823, Mar. 2021.

[6] N. Afzal et al., "Submillimeter lateral displacement enables friction sensing and awareness of surface slipperiness," *IEEE Trans. Haptics*, vol. 15, no. 1, pp. 20–25, Jan./Mar. 2022.

[7] T. Andre, V. Levesque, V. Hayward, P. Lefevre, and J. L. Thonnard, "Effect of skin hydration on the dynamics of fingertip gripping contact," *J. Roy. Soc. Interface*, vol. 8, no. 64, pp. 1574–1583, Nov. 2011.

[8] G. Corniani, M. A. Casal, S. Panzeri, and H. P. Saal, "Population coding strategies in human tactile afferents," *PLoS Comput. Biol.*, vol. 18, no. 12, pp. 1–24, Dec. 2022.

[9] H. P. Saal, I. Birznieks, and R. S. Johansson, "Memory at your fingertips: How viscoelasticity affects tactile neuron signaling," *Elife*, vol. 12, 2023, Art. no. RP89616.

[10] L. Willemet, H. Nicolas, and W. Michaël. "Efficient tactile encoding of object slippage," *Sci. Rep.*, vol. 12, no. 1, p. 13192, Aug. 2022.

[11] W. Chen, H. Khamis, I. Birznieks, N. F. Lepora, and S. J. Redmond, "Tactile sensors for friction estimation and incipient slip detection—Toward dexterous robotic manipulation: A review," *IEEE Sensors J.*, vol. 18, no. 22, pp. 9049–9064, Nov. 2018.

[12] X. Song, H. Liu, K. Althoefer, T. Nanayakkara, and L. D. Seneviratne, "Efficient break-away friction ratio and slip prediction based on haptic surface exploration," *IEEE Trans. Robot.*, vol. 30, no. 1, pp. 203-219, Feb. 2014.

[13] J. W. James and N. F. Lepora, "Slip detection for grasp stabilization with a multifingered tactile robot hand," *IEEE Trans. Robot.*, vol. 37, no. 2, pp. 506–519, Apr. 2021.

[14] S. Cui, S. Wang, R. Wang, S. Zhang, and C. Zhang, "Learning-based slip detection for dexterous manipulation using GelStereo sensing," *IEEE Trans. Neural Netw. Learn. Syst.*, pp. 1-10, 2023.

[15] Y. Zhou et al., "T-TD3: A reinforcement learning framework for stable grasping of deformable objects using tactile prior," *IEEE Trans. Autom. Sci. Eng.*, early access, Aug., 2024, doi: 10.1109/TASE.2024.3440047.

[16] R. Sui, L. Zhang, Q. Huang, T. Li, and Y. Jiang, "A novel incipient slip degree evaluation method and its application in adaptive control of grasping force," *IEEE Trans. Autom. Sci. Eng.*, vol. 21, no. 3, pp. 2454-2468, July 2024.

[17] T. Kim, J. Kim, I. You, J. Oh, S.-P. Kim, and U. Jeong, "Dynamic tactility by position-encoded spike spectrum," *Sci. Robot.*, vol. 7, no. 63, Feb. 2022, Art. no. eabl5761.

[18] Q. Wang, P. M. Ulloa, R. Burke, D. C. Bulens and S. J. Redmond, "Robust learning-based incipient slip detection using the PapillArray optical tactile sensor for improved robotic gripping," *IEEE Robot. Automat. Lett.*, vol. 9, no. 2, pp. 1827-1834, Feb. 2024.

[19] J. Yu, S. Yao, X. Li, A. Ghaffar and Z. Yao, "Design of a 3-D tactile sensing array for incipient slip detection in robotic dexterous manipulation," *IEEE Trans. Instrum. Meas.*, vol. 73, pp. 1-14, 2024, Art no. 9514214.

[20] H. Zhang, J. Long, X. Kong, and P. Yu, "Localized displacement phenomenon of a sliding soft fingertip under different grasp force for slip prediction on prosthetic hand," *Measurement*, vol. 194, 2022, Art. no. 111092.

[21] S. Zhang et al., "Artificial skin based on visuo-tactile sensing for 3d shape reconstruction: Material, method, and evaluation," *Adv. Funct. Mater.*, 2024, Art. no. 2411686.

[22] M. Li, T. Li, and Y. Jiang, "Marker displacement method used in vision-based tactile sensors—From 2-D to 3-D: A review," *IEEE Sensors J.*, vol. 23, no. 8, pp. 8042–8059, Apr. 2023.

[23] S. Dong, D. Ma, E. Donlon, and A. Rodriguez, "Maintaining grasps within slipping bounds by monitoring incipient slip," in *Proc. Int. Conf. Robot. Autom. (ICRA)*, May 2019, pp. 3818–3824.

[24] D. C. Bulens et al., "Incipient slip detection with a biomimetic skin morphology," in *Proc. IEEE/RSJ Int. Conf. Intell. Robots Syst. (IROS)*, Oct. 2023, pp. 8972-8978.

[25] D. -J. Boonstra, L. Willemet, J. Luijkx and M. Wiertlewski, "Learning to estimate incipient slip with tactile sensing to gently grasp objects," in *Proc. Int. Conf. Robot. Autom. (ICRA)*, 2024, pp. 16118-16124.

[26] N. Zhang, R. Sui, L. Zhang, T. Li, and Y. Jiang, "A robust incipient slip detection method with vision-based tactile sensor based on local deformation degree," *IEEE Sensors J.*, vol. 23, no. 15, pp. 17200-17213, Aug. 2023.

[27] R. Sui, L. Zhang, T. Li, and Y. Jiang, "Incipient slip detection method for soft objects with vision-based tactile sensor," *Measurement*, vol. 203, Nov. 2022, Art. no. 111906.

[28] M. Li, Y. H. Zhou, T. Li, and Y. Jiang, "Incipient slip-based rotation measurement via visuotactile sensing during in-hand object pivoting," in *Proc. IEEE Int. Conf. Robot. Autom. (ICRA)*, 2024, pp. 17132-17138.

[29] A. Parag, E. H. Adelson and E. Misimi, "Learning incipient slip with GelSight sensors: Attention Classification with Video Vision Transformers," in *Proc. IEEE/RSJ Int. Conf. Intell. Robots Syst. (IROS)*, Oct. 2024, pp. 13960-13966.

[30] J. Lu, B. Niu, H. Ma, J. Zhu and J. Ji, "STNet: Spatio-temporal fusion-based self-attention for slip detection in visuo-tactile sensors," in *Proc. IEEE Int. Conf. Robot. Autom. (ICRA)*, 2024, pp. 3051-3056.

[31] J. L. Johnson, Contact Mechanics. Cambridge, U.K., MA, USA: Cambridge Univ. Press, 1987.

[32] V. L. Popov, Contact mechanics and friction. Berlin, Springer Berlin Heidelberg, 2010.

[33] B. P. Delhaye, A. Barrea, B. B. Edin, P. Lefèevre, and J. L. Thonnard, "Surface strain measurements of fingertip skin under shearing," *J. Roy. Soc. Interface*, vol. 13, 2016, Art. no. 20150874.

[34] B. P. Delhaye, E. Jarocka, A. Barrea, J. L. Thonnard, B. B. Edin, and P. Lefèevre, "High-resolution imaging of skin deformation shows that afferents from human fingertips signal slip onset," *Elife*, vol. 10, 2021, Art. no. e64679.

[35] L. Willemet, K. Kanzari, J. Monnoyer, I. Birznieks, and M. Wiertlewski, "Initial contact shapes the perception of friction," *Proc. Nat. Acad. Sci. USA*, vol. 118, no. 49, Dec. 2021.

[36] S. du Bois de Dunilac, D. C. Bulens, P. Lefèvre, S. J. Redmond, and B. P. Delhaye, "Biomechanics of the finger pad in response to torsion," *J. Roy. Soc. Interface*, vol. 20, no. 201, 2023, Art. no. 20220809.

[37] Y. Li et al., "Imaging dynamic three-dimensional traction stresses," *Sci. Adv.*, vol. 8, no. 11, Mar. 2022, Art. no. eabm0984.

[38] J Ding, H. Wu, J. Han, and S. Yan, "Analysis of stress drop behavior for stick-slip friction based on total reflection observation experiment," *Tribol. Int.*, vol. 192, Apr. 2024, Art. no. 109272.

[39] X. Shi, "On slip inception and static friction for smooth dry contact," *ASME J. Appl. Mech.*, vol. 81, p. 121005, 2014.

[40] Y. Shao, V. Hayward, and Y. Visell, "Compression of dynamic tactile information in the human hand," *Sci. Adv.*, vol. 6, no. 16, 2020, Art. no. eaaz1158.

[41] V. A. Ho and S. Hirai, Mechanics of localized slippage in tactile sensing. New York, NY, USA: Springer, 2014.

[42] J. Van Brakel. "Robust peak detection algorithm using Z-scores". [Online]. Available: https://stackoverflow.com/questions/22583391/peak-signal-detection-in-realtime-timeseries-data

[43] M. Li, L. Zhang, T. Li and Y. Jiang, "EasyCalib: Simple and low-cost in-situ calibration for force reconstruction with vision-based tactile sensors," *IEEE Robot. Autom. Lett.*, vol. 9, no. 9, pp. 7803-7810, 2024.

[44] L. Zhang, T. Li, and Y. Jiang, "Improving the force reconstruction performance of vision-based tactile sensors by optimizing the elastic body," *IEEE Robot. Autom. Lett.*, vol. 8, no. 2, pp. 1109–1116, Feb. 2023.

[45] Q. K. Luu et al., "Simulation, learning, and application of vision-based tactile sensing at large scale," IEEE Trans. Robot., vol. 39, no. 3, pp. 2003–2019, Jun. 2023.

[46] C. Zhao, J. Liu and D. Ma, "iFEM2.0: Dense 3-D contact force field reconstruction and assessment for vision-based tactile sensors," *IEEE Trans. Robot.*, vol. 41, pp. 289-305, 2025.

[47] Z. Liu and R. D. Howe, "Beyond Coulomb: Stochastic friction models for practical grasping and manipulation," *IEEE Robot. Autom. Lett.*, vol. 8, no. 8, pp. 5140-5147, Aug. 2023.

[48] M. Li, L. Zhang, T. Li, and Y. Jiang, "Learning gentle grasping from human-free force control demonstration," *IEEE Robot. Autom. Lett.*, vol. 10, no. 3, pp. 2391-2398, March 2025.